\documentclass[10pt,twocolumn,letterpaper]{article}

\usepackage{iccv}              %

\definecolor{iccvblue}{rgb}{0.21,0.49,0.74}
\usepackage[pagebackref,breaklinks,colorlinks,allcolors=iccvblue]{hyperref}

\def\confName{ICCV}
\def\confYear{2025}

\usepackage[capitalize]{cleveref}
\crefname{section}{Sec.}{Secs.}
\Crefname{section}{Section}{Sections}
\Crefname{table}{Table}{Tables}
\crefname{table}{Tab.}{Tabs.}

\usepackage{adjustbox}
\usepackage{threeparttable}
\usepackage{xcolor,colortbl}
\usepackage{multirow}
\usepackage{makecell}
\usepackage{siunitx}
\usepackage{tcolorbox}
\usepackage[accsupp]{axessibility} %
\usepackage{microtype}
\usepackage{marvosym}
\usepackage{tcolorbox}

\newcommand{\ourmodel}{{{DocThinker}}\xspace}
\newcommand\mypara[1]{\vspace{1.0mm}\noindent\textbf{#1}}

\def\vs{{\em vs.~}}

\definecolor{blue2}{RGB}{20, 54, 254}

\definecolor{deepblue}{RGB}{59,113,170}
\definecolor{contentwhite}{RGB}{255,255,255}

\definecolor{lightgrey}{gray}{0.93}

\definecolor{red2}{RGB}{252, 54, 65}

\usepackage[accsupp]{axessibility}  %

\title{DocThinker: Explainable Multimodal Large Language Models with Rule-based Reinforcement Learning for Document Understanding}

\author{
  Wenwen Yu$^{1}$,
  Zhibo Yang$^{2}$,
  Yuliang Liu$^{1}$, 
  Xiang Bai$^{1}$\textsuperscript{\Letter}       \\
    \small{\textsuperscript{\rm 1}Huazhong University of Science and Technology}, \small{\textsuperscript{\rm 2}Alibaba Group} \\
    {\scriptsize\texttt{\{wenwenyu,ylliu,xbai\}@hust.edu.cn},} {\scriptsize\texttt{yangzhibo450@gmail.com}}
  }

\begin{document}
\maketitle

\def\thefootnote{\Letter}\footnotetext{Corresponding author.}

\begin{abstract}

Multimodal Large Language Models (MLLMs) have demonstrated remarkable capabilities in document understanding. However, their reasoning processes remain largely black-box, making it difficult to ensure reliability and trustworthiness, especially in high-stakes domains such as legal, financial, and medical document analysis. Existing methods use fixed Chain-of-Thought (CoT) reasoning with supervised fine-tuning (SFT) but suffer from catastrophic forgetting, poor adaptability, and limited generalization across domain tasks.
In this paper, we propose \ourmodel, a rule-based Reinforcement Learning (RL) framework for dynamic inference-time reasoning. Instead of relying on static CoT templates, \ourmodel autonomously refines reasoning strategies via policy learning, generating explainable intermediate results, including structured reasoning processes, rephrased questions, regions of interest (RoI) supporting the answer, and the final answer. By integrating multi-objective rule-based rewards and KL-constrained optimization, our method mitigates catastrophic forgetting and enhances both adaptability and transparency.
Extensive experiments on multiple benchmarks demonstrate that \ourmodel significantly improves generalization while producing more explainable and human-understandable reasoning steps. Our findings highlight RL as a powerful alternative for enhancing explainability and adaptability in MLLM-based document understanding. Code will be available at https://github.com/wenwenyu/DocThinker.

\end{abstract}

\section{Introduction}
\label{sec:intro}

Multimodal Large Language Models (MLLMs)~\cite{ChatGPT, GPT-4, GPT-4V(ision)} have significantly advanced document understanding, yet their reasoning mechanisms remain largely opaque. This lack of explainability~\cite{Hsieh2024ACG,Sun2024ARO} limits their application in high-stakes domains such as legal, financial, and medical document analysis, where transparency is critical for ensuring trustworthiness. Unlike human reasoning, which involves structured and multi-step inference, MLLMs typically operate as black-box systems, making it difficult to validate their decision-making process~\cite{dang2024explainable}. While Chain-of-Thought (CoT)~\cite{Wei2022cot} prompting has been widely adopted to enhance explainability, existing approaches heavily rely on static reasoning templates, which struggle to generalize across diverse complex scenarios and tasks.

\begin{figure}[tbp]
    \centering 
    \centerline{\includegraphics[width=0.98\linewidth]{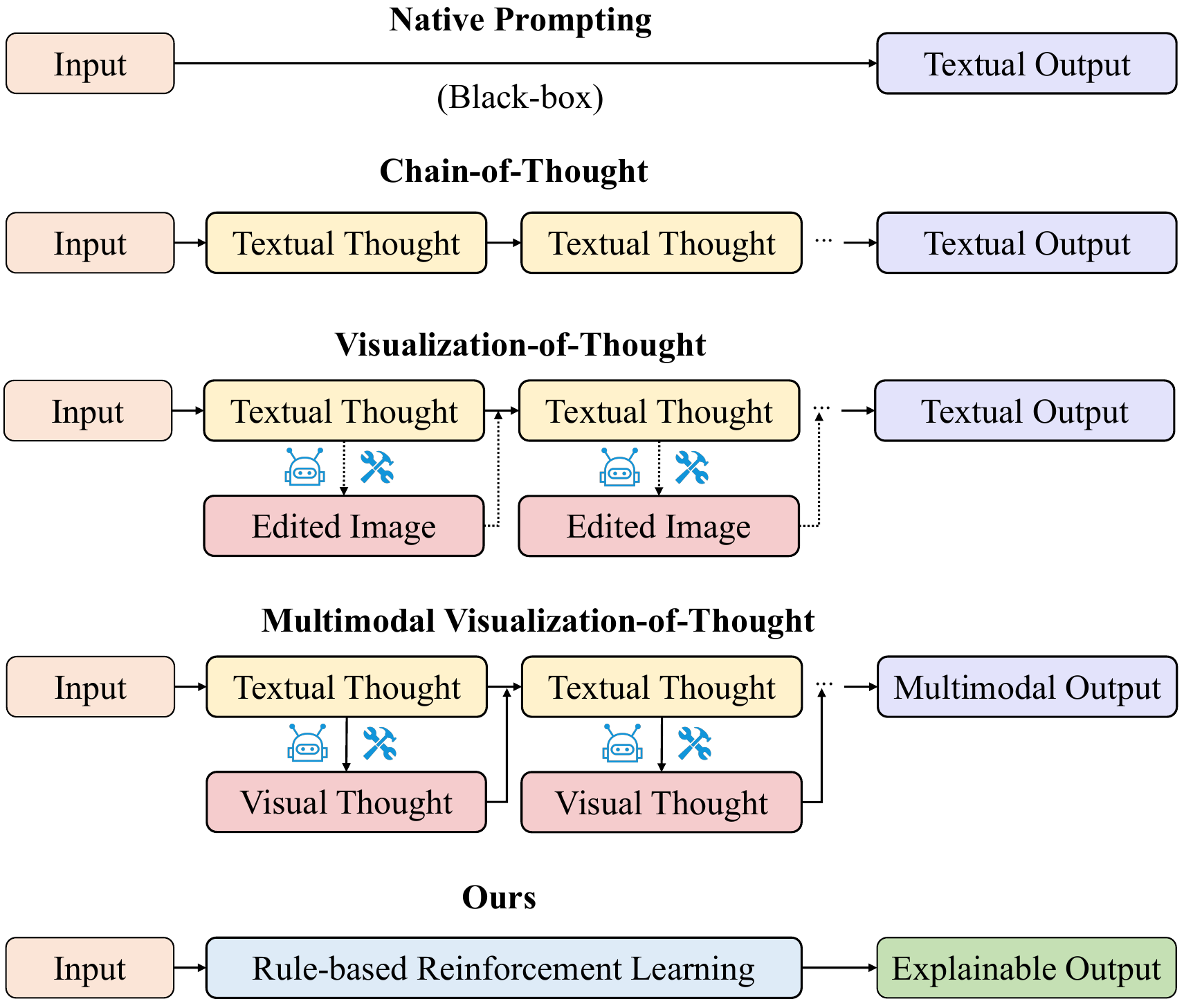}}
    \vspace{-2mm}
    \caption{\textbf{Comparison of different approaches for improving model's explainability and transparency in MLLM-based document understanding.} Traditional methods, including Chain-of-Thought (CoT), Visualization-of-Thought (VoT), and Multimodal Visualization-of-Thought (MVoT) rely on static reasoning templates, predefined heuristics, or external agents and tools, limiting their adaptability and generalization. In contrast, the proposed \ourmodel leverages rule-based reinforcement learning to explore diverse reasoning paths and generate explainable intermediate steps, including reasoning traces, rephrased questions, regions of interest (RoI) supporting the answer, and the final answer, enabling more adaptive and explainable document understanding.}
    \label{fig:intropipeline}
    \vspace{-6mm}
\end{figure}

To address this, recent research has explored multimodal CoT reasoning techniques, as shown in~\cref{fig:intropipeline}. ReFocus~\cite{Fu2025ReFocusVE} presents a visual-editing-based CoT framework, allowing models to selectively highlight and modify key regions via invoking external tools and agents in structured document images, improving comprehension of charts and tables. Visual CoT~\cite{Shao2024viscot} introduces multi-turn processing pipelines that dynamically focus on key regions in visual images, enabling more explainable intermediate reasoning steps. Similarly, Multimodal Visualization-of-Thought (MVoT)~\cite{Li2025mvot} extends CoT reasoning by generating interleaved visual and textual reasoning traces, aiming to improve transparency. The Mind’s Eye of LLMs~\cite{Wu2024vot} further proposes Visualization-of-Thought (VoT), a technique that elicits spatial reasoning by generating visual representations of thought processes. But this method applies in navigation-based applications, and its adaptability to document understanding remains limited. Besides, these approaches still depend on predefined heuristics and static reasoning paths, making them inherently inflexible and susceptible to catastrophic forgetting and poor generalization across varied document types and tasks.

Another emerging direction is reinforcement learning (RL)-based~\cite{Sutton1998IntroductionTR} reasoning, which has shown promise in overcoming the rigidity of fixed CoT methods. DeepSeek-R1~\cite{DeepSeekAI2025DeepSeekR1IR} framework demonstrates that pure RL training can incentivize emergent reasoning behaviors without relying on extensive supervised fine-tuning (SFT), achieving state-of-the-art performance in complex reasoning tasks. Inspired by this, MedVLM-R1~\cite{Pan2025MedVLMR1IM} applies RL techniques to medical vision-language models, proving its effectiveness in enhancing transparency and generalization in medical image understanding. Visual-RFT~\cite{Liu2025VisualRFTVR} introduces Visual Reinforcement Fine-Tuning, a reward-driven optimization framework designed to enhance the reasoning capabilities of vision-language models. Unlike conventional supervised fine-tuning, which relies on large annotated datasets, Visual-RFT employs verifiable reward functions to guide learning, significantly improving data efficiency. Experimental results indicate that reinforcement fine-tuning improves performance in open-vocabulary detection, few-shot object recognition, and reasoning grounding tasks, demonstrating its ability to generalize across diverse visual domains. While these methods primarily focus on general visual tasks, their success highlights the potential of RL in optimizing reasoning strategies for MLLMs. However, RL-based approaches for document understanding remain underexplored, particularly in designing effective reward functions that optimize both reasoning adaptability and explainability.

While humans naturally employ structured, multi-step reasoning when interpreting documents, integrating inference-time reasoning into MLLM-based document understanding is still an open challenge. Inspired by recent advancements, we propose \ourmodel, a novel rule-based Reinforcement Learning (RL) framework designed for inference-time reasoning in document understanding. Unlike fixed CoT or VoT-style methods, \ourmodel explores diverse reasoning paths and get explainable intermediate steps, including explicit reasoning traces, rephrased questions, regions of interest (RoI) supporting the answer, and the final answer, highlighting its ability to produce more flexible and varied outputs compared to traditional CoT. Instead of following fixed reasoning templates, \ourmodel autonomously refines its reasoning process through policy learning based on the Group Relative Policy Optimization (GRPO) algorithm~\cite{Shao2024DeepSeekMathPT}, mitigating catastrophic forgetting and improving adaptability. The model is trained using reinforcement learning with a proposed multi-objective reward function, enabling it to self-adapt to diverse document structures while preserving explainability. Although VoT-like methods generate grounded thought, they lack revision ability. Our RL enables self-reflection and correction that is complementary to VoT-like methods. Additionally, KL-constrained optimization is employed to ensure stable policy updates, preventing reward exploitation and preserving reasoning coherence.

Our main contributions are summarized as follows:
\begin{itemize}

\item  We introduce \ourmodel, to the best of our knowledge, the first RL-based framework for document understanding, enabling adaptive inference-time reasoning without relying on fixed CoT templates.

\item We propose a set of multi-objective reward functions that incentivize the model to generate human-understandable reasoning steps, while ensuring robust generalization across diverse document types and tasks.

\item We conduct extensive experiments on multiple benchmark datasets, demonstrating that \ourmodel significantly improves generalization and explainability compared to existing CoT-based and SFT-based methods. Our findings highlight the potential of RL as a key enabler for more explainable, adaptable, and reliable MLLM-based document understanding systems.

\end{itemize}

\section{Related Work}
\label{sec:related}

\subsection{MLLMs for Document Understanding}

Multimodal Large Language Models (MLLMs) have shown strong potential in document understanding by integrating textual and visual elements. Existing approaches enhance comprehension through layout-awareness, high-resolution processing, and specialized encoding techniques. DocLLM~\cite{Wang2023DocLLMAL}, LLaVAR~\cite{zhang2023llavar}, and mPLUG-DocOwl~\cite{Ye2023mPLUGDocOwlMM} improve text-centric document reasoning via instruction tuning, while methods like DocPedia~\cite{feng2024docpedia} and Vary~\cite{wei2024vary} refine image processing for structured text extraction. Other models, such as UReader~\cite{ye2023ureader} and InternVL1.5~\cite{chen2024far}, incorporate OCR and adaptive resolution techniques to enhance text recognition.
To further optimize efficiency, models like TextMonkey~\cite{Liu2024TextMonkeyAO} and Fox~\cite{liu2024focus} introduce token compression and unified encoding for multi-page document analysis. Additional improvements, including compression strategies in DocKylin~\cite{Zhang2024DocKylinAL} and multi-scale integration in StrucTexTv3~\cite{Lyu2024StrucTexTv3AE}, enhance structure-aware reasoning.
Despite these advancements, explainability remains a critical limitation. While step-wise reasoning frameworks~\cite{Zhang2024ReadAT} and visual editing-based approaches like ReFocus~\cite{Fu2025ReFocusVE} improve transparency, they rely on static reasoning strategies that hinder adaptability. Most models depend on supervised fine-tuning (SFT), which improves task performance but often leads to overfitting and weak generalization across diverse document types. This highlights the need for adaptive learning frameworks capable of dynamic reasoning refinement while preserving explainability across varied document scenarios.

\subsection{RL for Explainability and Reasoning}

Explainability is crucial for deploying MLLMs in sensitive domains, ensuring transparency and trust in model decisions~\cite{dang2024explainable,Sun2024ARO}. Reinforcement Learning (RL) has emerged as a promising alternative to supervised fine-tuning (SFT)~\cite{Chu2025SFTMR}, addressing overfitting and limited generalization by allowing models to self-improve reasoning strategies through interaction with an environment. Unlike static Chain-of-Thought (CoT) approaches, RL enhances adaptability, explainability, and generalization in complex reasoning tasks.  
Recent advancements demonstrate RL’s effectiveness in language and vision-language models. The OpenAI o1 model~\cite{jaech2024openai} applies RL to enhance reasoning capabilities, while DeepSeek-R1-Zero~\cite{DeepSeekAI2025DeepSeekR1IR} achieves better reasoning and thinking process ability by training entirely with RL, incentivizing emergent reasoning via its Group Relative Policy Optimization (GRPO) framework without relying on SFT. MedVLM-R1~\cite{Pan2025MedVLMR1IM} extends this to medical image analysis, showing improved explainability and transparency.  
In multimodal learning, Visual-RFT~\cite{Liu2025VisualRFTVR} introduces verifiable reward functions, improving data efficiency and reasoning adaptability in open-vocabulary detection, few-shot recognition, and grounding tasks. RLHF-V~\cite{Yu2023RLHFVTT} further aligns MLLMs with human trustworthiness through fine-grained RL-based feedback.  
Despite RL’s success in vision-language tasks, its application in document understanding remains largely unexplored. Existing RL-based approaches fail to jointly process text, layout, and visual elements while maintaining explainability. Furthermore, designing effective reward functions that balance reasoning adaptability and explainability remains an open challenge.  
To bridge this gap, we propose \ourmodel, a rule-based RL framework optimized with Group Relative Policy Optimization (GRPO)~\cite{Shao2024DeepSeekMathPT}, incorporating verifiable multi-objective rewards to enable inference-time reasoning for complex document understanding.

\begin{figure*}[htbp]
    \centering
    \centerline{\includegraphics[width=1.0\linewidth]{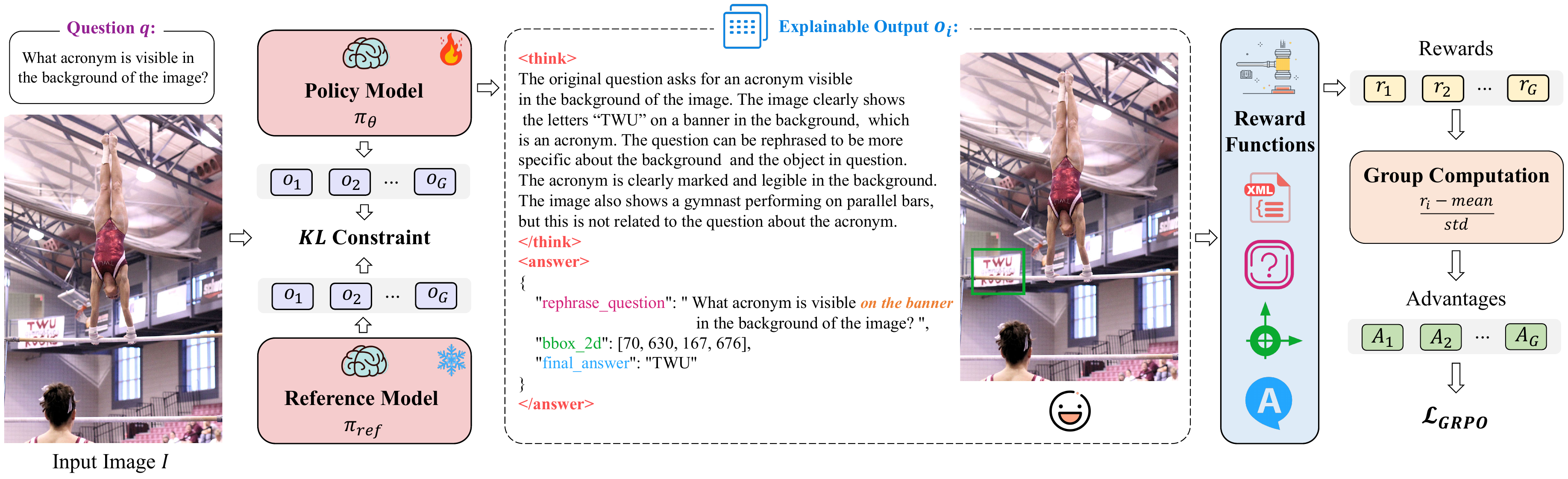}}
    \captionsetup{width=1.0\textwidth}
    \vspace{-2mm}
    \caption{\textbf{Schematic illustration of the proposed \ourmodel framework.} Given an input image $I$ and question $q$, we first sample $G$ candidate outputs $\left\{o_i\right\}_{i=1}^G$ from the old policy model $\pi_{\theta_{old}}$. The $i$-th output $o_i$ has explainable and human-understandable intermediate results, including reasoning processes, rephrased question, regions of interest (RoI) supporting the answer, and final answer. Then we compute a reward $r_i$ for each output $o_i$ using our proposed multi-objective reward functions, which will be detailed in \cref{sec:multi_obj_reward}, including XML format reward, rephrased question reward, RoI IoU reward, and final answer accuracy reward. Subsequently, each reward $r_i$ is normalized by subtracting the group average $mean$ and dividing by the group standard deviation $std$ to get a group relative advantage $A_i$. Finally, we optimize the current policy model $\pi_{\theta}$ where $\theta$ is trainable parameters by maximizing the advantage $A_i$ while ensuring that the model remains close to the reference policy model $\pi_{ref}$, via KL divergence between $\pi_{\theta}$ and $\pi_{ref}$.}
    \label{fig:diagram}
    \vspace{-6mm}
\end{figure*}

\section{Methodology}\label{sec:method}

\subsection{Preliminary}

\paragraph{Group Relative Policy Optimization (GRPO).}
The GRPO algorithm, first introduced in DeepSeekMath~\cite{Shao2024DeepSeekMathPT}, is a reinforcement learning framework designed to improve reasoning without the need for a separate critic model, a key limitation of existing methods such as Proximal Policy Optimization (PPO)\cite{PPO}. Traditional RL approaches like PPO rely on a value network to estimate the quality of model predictions, which can introduce instability and additional computational costs. In contrast, GRPO directly compares a group of generated responses, making it a more efficient alternative for large-scale language model training.

In GRPO, given a question $q$, the old policy model $\pi_{\theta_{\text{old}}}$ first generates a group of different candidate response outputs $\{o_1, o_2, ..., o_G\}$ with size of $G$. These response outputs are then evaluated through a rule-based reward function $R(q, o)$ to obtain $G$ rewards denoted as $\{r_1, r_2, ..., r_G\}$ correspondingly, which is defined as follows:
\begin{equation}
    r_i = R(q, o_i) = 
    \begin{cases} 
    1 ,& \text{if } o_i \text{ = ground truth}, \\
    0 ,& \text{otherwise},
    \end{cases} 
\end{equation}
where $R(\cdot, \cdot)$ is the rule-based verifiable reward function. $R$ takes the question and output pair $(q,o_i)$ as inputs, and checks whether the prediction $o_i$ is correct compared to ground truth under predefined rules. In our works, we proposed multi-objective reward functions tailored for document understanding, which will be detailed in \cref{sec:multi_obj_reward}, to incentivize the model to generate human-understandable reasoning steps, while ensuring robust generalization across diverse document types and tasks.

Instead of computing absolute values for each response, GRPO normalizes the rewards within the group, ensuring that the model learns from relative advantages. Specifically, the advantage is computed by taking the difference between each reward and the $mean$ of the group, normalized by the standard deviation $std$, formulated as follows: 
\begin{align}
A_i &= \frac{r_i - \text{mean}(\{r_1, \dots, r_G\})}{\text{std}(\{r_1, \dots, r_G\})},
\end{align}
where $A_i$ represents the advantage of $i$-th output $o_i$, meaning the relative quality of the $i$-th responses. The advantage $A_i$ is sequence-level normalized reward, and we set the advantage $A_{i,t}$ of $t$-th auto-regressive decoding time step token in the output $o_i$ as the sequence-level advantage $A_i$. This process eliminates the need for a critic network, making policy updates computationally efficient and stable. The intuition behind GRPO objective is to maximize the advantage of the generated responses, while ensuring that the model remains close to the reference policy model $\pi_{{ref}}$. Consequently, the GRPO loss $\mathcal{L}_{\text{GRPO}}$ is defined as follows: 
\begin{align}
\mathcal{L}_{\text{GRPO}}(\theta) &= -\frac{1}{G}\sum_{i=1}^G \frac{1}{\left|o_i\right|} \sum_{t=1}^{\left|o_i\right|}  \bigg[  
\frac{\pi_{\theta}(o_{i,t} \mid q, o_{i,\textless t})}{ \varphi\big[\pi_{\theta}(o_{i,t} \mid q, o_{i,\textless t} ) \big] } A_{i,t}  \notag \\
&\quad - \beta \mathbb{D}_{\text{KL}}(\pi_{\theta} \parallel \pi_{\text{ref}}) \bigg] ,
\label{eq:simply_grpo}
\end{align}
where the first term represents the scaled advantage and the second term is regularization to penalize deviations from the reference policy $\pi_{ref}$ through Kullback–Leibler (KL) divergence $\mathbb{D}_{\text{KL}}(\pi_{\theta} \parallel \pi_{\text{ref}})$, helping prevent catastrophic forgetting. $\varphi[\cdot]$ represents stop gradient operation. $\theta$ is the trainable parameter of the current policy model $\pi_{\theta}$. $\beta \in \mathbb{R} \geq 0$ is a hyper-parameter and controls the regularization strengths. For more introduction of the general version of GRPO, please refer to appendix.%

\subsection{\ourmodel}
As illustrated in \cref{fig:diagram}, we introduce \ourmodel, a reinforcement learning (RL)-based framework designed for inference-time reasoning in multimodal document understanding. Unlike conventional supervised fine-tuning (SFT), which is often prone to overfitting and limited generalization, \ourmodel refines its reasoning strategies by leveraging rule-based reward signals, enabling greater robustness across diverse document types.

Built on Group Relative Policy Optimization (GRPO), \ourmodel optimizes for explainability, adaptability, and accuracy, allowing the model to iteratively improve its decision-making process. While the original GRPO algorithm has been primarily applied to text-only question-answering tasks, \ourmodel extends its application to multimodal settings, where both document images and textual queries serve as inputs.
Given a document image $I$ and a question $q$, \ourmodel generates a response $o_i$ that includes explainable intermediate reasoning results, such as explicit reasoning traces, rephrased questions for improved clarity, identified Regions of Interest (RoI) supporting the answer, and the final predicted response. 
Through a multi-objective reward system, \ourmodel continuously refines its reasoning strategy via the GRPO algorithm, ensuring stable learning and enhanced adaptability across a wide range of document scenarios. 
The following sections discuss the choice of base model and prompt template in \cref{sec:basemodel}, and the design of verifiable multi-objective rewards for document understanding tasks in \cref{sec:multi_obj_reward}.

\subsubsection{Base Model and Prompt Template}\label{sec:basemodel}
For our base model, we adopt the state-of-the-art multimodal large language model Qwen2.5-VL 3B and 7B~\cite{Shuai2025Qwen2.5-VL}, denoted as $\pi_{\theta}$, where $\theta$ are the trainable parameters. Given a training dataset $X$, each sample $x$ consists of a document image $I$ and a text prompt $p$, which includes the user’s question $q$ alongside a fixed template message. The prompt template is designed to instruct the MLLM to produce structured output ${o}$, which includes both a reasoning trace and a final output encoded in designated XML-like tags (\textless think\textgreater...\textless /think\textgreater and \textless answer\textgreater...\textless /answer\textgreater). 
The reasoning trace enclosed in \textless think\textgreater...\textless /think\textgreater serves as a key component in the model’s self-improvement and optimization process during reinforcement fine-tuning. The answer enclosed in \textless answer\textgreater...\textless /answer\textgreater is formatted in JSON, containing three critical fields: ``rephrase\_question'', ``bbox\_2d'', and ``final\_answer''.
The ``rephrase\_question" represents an improved, more descriptive version of the original query. This refinement reduces ambiguity, helping users better understand how the model processes the question and formulates an inference. By enhancing question clarity, this component significantly contributes to the model’s overall explainability.
The ``bbox\_2d" encodes the 2D bounding box coordinates corresponding to the regions of the document image that the model deems highly relevant to answering the question. This visual cue serves as an explainability aid, providing insights into which parts of the document influence the model’s reasoning.
The ``final\_answer" contains the model’s predicted response to the given question $q$, ensuring that all intermediate reasoning steps contribute to a well-supported final decision.
By incorporating both textual and visual reasoning elements, \ourmodel enhances explainability across two modalities. The ``rephrase\_question" field refines the linguistic aspect of reasoning, while the ``bbox\_2d" field introduces a visual grounding mechanism, making the model’s decision-making process more transparent. For further details of the prompt template, please refer to appendix.

\mypara{Optimization.} We adopt the GRPO-based RL objective formulated in \cref{eq:simply_grpo} to optimize $\pi_{\theta}$, ensuring that the generated answers are accurate, well-structured, and transparently reasoned. The reasoning trace, enclosed within \textless think\textgreater...\textless /think\textgreater, serves as a crucial component for self-learning, allowing the model to iteratively refine its reasoning process. Meanwhile, the reward signal, derived from the correctness of the final answer, guides policy optimization, reinforcing high-quality responses. Through this structured reinforcement learning framework, \ourmodel achieves greater explainability and generalization in multimodal document understanding.

\subsubsection{Multi-objective Reward Functions}\label{sec:multi_obj_reward}
The reward model plays a crucial role in reinforcement learning (RL) by aligning the model’s predictions with predefined correctness criteria. While traditional RL approaches often rely on human preference-based reward models\cite{liu2024skywork,zang2025internlm}, recent advancements, such as DeepSeek-R1~\cite{DeepSeekAI2025DeepSeekR1IR}, have demonstrated that verifiable reward functions can significantly enhance a model’s reasoning ability. Inspired by this success, we extend Reinforcement Learning with Verifiable Rewards (RLVR) to the visual document understanding domain by designing a rule-based multi-objective reward function that evaluates both textual reasoning and visual comprehension. This ensures that the model not only produces accurate answers but also generates explainable intermediate steps, improving both generalization and transparency.
Our framework evaluates model outputs based on four core criteria: format compliance, final answer accuracy, region of interest (RoI) consistency, and rephrased question quality.

\mypara{Format Reward.}
The format reward ensures that the model's output adheres to a structured XML-style schema, enforcing consistency for explainable and machine-parsable outputs. It verifies that the reasoning trace is enclosed in \textless think\textgreater...\textless /think\textgreater tags and that the final response in \textless answer\textgreater...\textless /answer\textgreater is valid JSON with required key-value pairs. Outputs deviating from this format are penalized to maintain structured and systematic reasoning.
Given a model output $o$, the format reward is defined as:
\[
R_{\text{format}} =
\begin{cases} 
1, & \begin{array}{l} 
\text{if $o$ follows the XML-style schema and} \\ 
\text{JSON structure with valid key-value pairs,} 
\end{array} \\[8pt]
0, & \text{otherwise.}
\end{cases}
\]

\mypara{Accuracy Reward.} 
The final answer accuracy reward measures whether the model’s generated response ``final\_answer'' aligns with the ground truth answer. Unlike traditional RLHF, where correctness is determined through human preference rankings, we leverage a direct verification function that compares the model’s final answer against predefined ground truth values.
For a given question $q$ and model-generated answer ``final\_answer'', the reward function is defined as:
\[
R_{\text{accuracy}} =
\begin{cases} 
1, & \text{if ``final\_answer'' = the ground truth,} \\
0, & \text{otherwise.}
\end{cases}
\]

\mypara{RoI IoU Reward.}
The RoI IoU reward evaluates how accurately the model identifies key visual regions in a document. 
For a given predicted bounding box $B_{\text{pred}} = $ ``bbox\_2d'' and ground truth bounding box \( B_{\text{gt}} \), the corresponding reward function is defined as:  
\[
R_{\text{RoI}} =
\begin{cases} 
1, & \text{if IoU}(B_{\text{pred}}, B_{\text{gt}}) \geq 0.5, \\
0, & \text{otherwise.}
\end{cases}
\]
This reward encourages precise localization of relevant document regions, ensuring extracted information directly supports the model’s final answer.

\begin{table*}[tbp]
\centering

 \begin{minipage}[b]{\textwidth}
 \resizebox{1\linewidth}{!}{
\begin{tabular}{l|c|c|c|c|c|c|c|c|c|c|c|c|c|c}
\toprule
  \multicolumn{4}{c|}{} & \multicolumn{6}{c}{{Document-oriented Understanding }} & \multicolumn{5}{c}{{General Multimodal Understanding }}  \\
\cmidrule(lr){1-4} \cmidrule(lr){5-10} \cmidrule(lr){11-15}
  \multirow{2}{*}{{MLLM}} & \multirow{2}{*}{{Res.}} & \multirow{2}{*}{{Data}} & \multirow{2}{*}{{Str.}} & \multicolumn{5}{c|}{{Doc/Text}}          & {Chart} & \multicolumn{2}{c|}{{General VQA}} & \multicolumn{3}{c}{{Relation Reasoning}}      \\
 \cmidrule(lr){5-9} \cmidrule(lr){10-0} \cmidrule(lr){11-12}\cmidrule(lr){13-15}  %
 &  &  &  & DocVQA & TextCaps & TextVQA &  DUDE &  SROIE & InfoQA  & F30k  & V7W & GQA      & OI      & VSR     \\
\midrule
 LLaVA-1.5-7B~\cite{liu2023improvedllava} &   336$^2$ & - & SFT & 0.244 & 0.597 & 0.588 & 0.290 &0.136 & 0.400  & 0.581  & \cellcolor{lightgrey}0.575 & 0.534 & 0.412 & 0.572      \\
 LLaVA-1.5-13B~\cite{liu2023improvedllava} &     336$^2$ & -  & SFT& 0.268 & 0.615 & 0.617 & 0.287 & 0.164 & 0.426   & 0.620  & \cellcolor{lightgrey}{0.580} &  0.571 & 0.413 & 0.590        \\
 SPHINX-13B~\cite{lin2023sphinx} &   224$^2$ & - & SFT & 0.198 & 0.551 & 0.532 & 0.000 & 0.071 & 0.352 & 0.607  & \cellcolor{lightgrey}0.558 &  0.584 & 0.467 & 0.613          \\
VisCoT-7B~\cite{Shao2024viscot} &   224$^2$ & 438k & SFT &      0.355    &   0.610  &   0.719   &  \cellcolor{lightgrey}0.279    &  \cellcolor{lightgrey}0.341   &  0.356    &    {0.671}    & \cellcolor{lightgrey}{0.580} & 0.616  &  \textbf{0.833}  &   {0.682}       \\
VisCoT-7B~\cite{Shao2024viscot}  &   336$^2$ & 438k & SFT  & {0.476} & {0.675} & {0.775} & \cellcolor{lightgrey}{0.386} & \cellcolor{lightgrey}{0.470} & 0.324  & 0.668  & \cellcolor{lightgrey}0.558 &  {0.631} & 0.822 & 0.614        \\

Qwen2.5VL-7B$^\dagger$~\cite{Shuai2025Qwen2.5-VL}  &   336$^2$  & - & - & \cellcolor{lightgrey}{0.350} & \cellcolor{lightgrey}{0.642} & \cellcolor{lightgrey}{0.735} & \cellcolor{lightgrey}{ 0.202 } & \cellcolor{lightgrey}{0.472} & \cellcolor{lightgrey}{0.325} &  \cellcolor{lightgrey} 0.603 & \cellcolor{lightgrey} 0.556 & \cellcolor{lightgrey} 0.455 & \cellcolor{lightgrey} 0.347 & \cellcolor{lightgrey} 0.616         \\
Qwen2.5VL-7B$^\dagger$~\cite{Shuai2025Qwen2.5-VL}  &   1536$^2$ & - & - &  \cellcolor{lightgrey}{}0.773  & \cellcolor{lightgrey}{}0.710 & \cellcolor{lightgrey}{}0.792 & \cellcolor{lightgrey}{}0.492 & \cellcolor{lightgrey}0.708 & \cellcolor{lightgrey}  0.663 & \cellcolor{lightgrey} 0.685 & \cellcolor{lightgrey} 0.604  & \cellcolor{lightgrey} 0.457 & \cellcolor{lightgrey}0.371 & \cellcolor{lightgrey}0.603      \\

Qwen2.5VL-7B*~\cite{Shuai2025Qwen2.5-VL}  &   336$^2$  & 4k & SFT & 0.355 & 0.658 & 0.740 & \cellcolor{lightgrey}{} 0.215 & \cellcolor{lightgrey}{} 0.489 & 0.334  & \cellcolor{lightgrey}{}0.624 & \cellcolor{lightgrey}{}0.563 & \cellcolor{lightgrey}{}0.467 & \cellcolor{lightgrey}{}0.405 & \cellcolor{lightgrey}{}0.619       \\
Qwen2.5VL-7B*~\cite{Shuai2025Qwen2.5-VL}  &   1536$^2$ & 4k & SFT & 0.784 & 0.725 & 0.801 & \cellcolor{lightgrey}{} 0.498 & \cellcolor{lightgrey}{}0.714  & 0.674  & \cellcolor{lightgrey}{}0.680& \cellcolor{lightgrey}{}0.609 & \cellcolor{lightgrey}{}0.472 & \cellcolor{lightgrey}{}0.427 & \cellcolor{lightgrey}{}0.624    \\

\midrule

\ourmodel-3B  &   336$^2$  & 4k & RL & 0.460 & 0.663 & 0.746 & \cellcolor{lightgrey}{}0.213  & \cellcolor{lightgrey}{}0.486 & 0.335 & \cellcolor{lightgrey}{} 0.664  & \cellcolor{lightgrey}{}0.572   & \cellcolor{lightgrey}{} 0.486   & \cellcolor{lightgrey}{}0.485   & \cellcolor{lightgrey}{}0.625           \\
\ourmodel-3B  &   1536$^2$  & 4k & RL & 0.751  & 0.691  & 0.762 & \cellcolor{lightgrey}{} 0.469 & \cellcolor{lightgrey}{}0.735 & 0.566 & \cellcolor{lightgrey}{}0.682  & \cellcolor{lightgrey}{}0.583  & \cellcolor{lightgrey}{}0.490  & \cellcolor{lightgrey}{}0.517  & \cellcolor{lightgrey}{} 0.637        \\
\ourmodel-7B  &   336$^2$  & 4k & RL & 0.579 & 0.682 & 0.802 & \cellcolor{lightgrey}{} 0.408 & \cellcolor{lightgrey}{} 0.495  &  0.347  & \cellcolor{lightgrey}{}  0.674 & \cellcolor{lightgrey}{} 0.580 & \cellcolor{lightgrey}{} 0.546 & \cellcolor{lightgrey}{}0.542 & \cellcolor{lightgrey}{}0.656        \\
\ourmodel-7B  &   1536$^2$ & 4k & RL & 0.795  & 0.738 & 0.827 & \cellcolor{lightgrey}{} 0.515 & \cellcolor{lightgrey}{} 0.806 &  0.689 & \cellcolor{lightgrey}{} 0.701 & \cellcolor{lightgrey}{}0.625 & \cellcolor{lightgrey}{}0.694 & \cellcolor{lightgrey}{}0.686 & \cellcolor{lightgrey}{}0.721        \\
\ourmodel-7B  &   1536$^2$ & 8k & RL & \textbf{0.802}  & \textbf{0.757} & \textbf{0.836} & \cellcolor{lightgrey}{} \textbf{0.568} & \cellcolor{lightgrey}{} \textbf{0.814} &  \textbf{0.697} & \textbf{0.734} & \cellcolor{lightgrey}{}\textbf{0.641} & \textbf{0.737} & 0.784 & \textbf{0.768}        \\

 \bottomrule
\end{tabular}
}
\end{minipage}
 \vspace{-2mm}
\caption{
\textbf{Performance on the Visual CoT benchmark.} \colorbox{lightgrey}{Grey} results indicate zero-shot performance. The final row uses 8k data including F30k, GQA, OI, and VSR; only DUDE, SROIE, V7W are zero-shot. Res. and Str. short for resolution and strategy. InfoQA, F30k, V7W, and OI short for InfographicsVQA, Flickr30k, Visual7W, and Open Images, respectively. $^\dagger$ indicate evaluating the model using the official checkpoint. * means trained it on 4data4k setting via supervised fine-tuning. DocThinker and Qwen2.5VL* differ only in training strategy. %
}
\label{tab:cot_benchmark}
\vspace{-6mm}
\end{table*}

\mypara{Rephrase Question Reward.}
To enhance explainability, we introduce a rephrase question reward, which evaluates how effectively the model reframes the original query for improved clarity. Since document-based questions are often ambiguous or underspecified, an ideal model should generate a well-structured and informative rephrased question that provides additional context without altering the intent.

The quality of the rephrased question is assessed based on two criteria: semantic similarity to the original query and word diversity. Given the original question $q_{\text{orig}} = q$ and the rephrased version $ q_{\text{rephrase}} =$ ``rephrase\_question'', we compute the soft reward, including cosine similarity $s$ and the ratio $r$ of new words compared to the original question:  
\[
R_{\text{rephrase}} =
\begin{cases} 
s+r, & \begin{array}{l} 
\text{if}  ~R_{\text{accuracy}} =1,
\end{array} \\[8pt]
0, & \text{otherwise,}
\end{cases}
\]
where $R_{\text{rephrase}}$ is normalized to [0, 1]. This reward encourages the model to generate refined queries that clarify ambiguous input while preserving the original meaning, ultimately improving transparency in its reasoning process.  

\mypara{Final Reward Function.}
The total reward combines four rewards to optimize both accuracy and explainability:
\[
R_{\text{total}} = \lambda_1 R_{\text{format}} + \lambda_2 R_{\text{accuracy}} + \lambda_3 R_{\text{RoI}} + \lambda_4 R_{\text{rephrase}},
\]
where hyperparameters $\lambda_i=1 $ balance reward contributions and avoid reward hacking. This joint optimization ensures the model generates structured, explainable, and verifiable reasoning outputs for document understanding.

\section{Experiments}

\mypara{Datasets.}
We utilize the Visual CoT dataset~\cite{Shao2024viscot} as training data, which contains 438k question-answer pairs annotated with intermediate bounding boxes that highlight key regions essential for answering questions. These bounding box annotations facilitate the computation of the RoI IoU reward during reinforcement learning, improving the model’s ability to focus on relevant areas within document images. The dataset spans five domains, including text/document understanding, fine-grained understanding, charts, general visual question answering (VQA), and relational reasoning.

We establish two training configurations to examine the model’s adaptability to different levels of data diversity. The 4data4k setup focuses on document understanding, selecting 1,000 samples each from DocVQA~\cite{Mathew2020DocVQAAD}, InfographicsVQA~\cite{Mathew2021InfographicVQA}, TextCaps~\cite{Sidorov2020TextCapsAD}, and TextVQA~\cite{Singh2019TowardsVMtextvqa} of Visual CoT dataset, totaling 4,000 instances. The 8data8k configuration extends training data to general VQA and relational reasoning by adding Flickr30k~\cite{Plummer2015Flickr30kEC}, GQA~\cite{Hudson2019GQAANgqa}, Open Images~\cite{Kuznetsova2018TheOIopenimage}, and VSR~\cite{Liu2022VisualSRvsr}, with 1,000 samples per dataset, totaling 8,000 instances. This comparison examines how broader domain coverage impacts generalization. Unless specified, we adopt 4data4k configuration as default setting. %

\mypara{Implementation Details.}
 Our base model is Qwen2.5-VL 3B and 7B~\cite{Shuai2025Qwen2.5-VL}, a state-of-the-art MLLM pretrained on curated web pages, open-source datasets, and synthetic data. To adapt it for document understanding tasks, we train it using the GRPO reinforcement learning framework, as described in \cref{sec:method}.
 The model is trained for two epochs on eight NVIDIA A100 80GB GPUs, with a batch size of 2. The number of generated candidate responses for GRPO is set to $G=6$. We employ the AdamW optimizer~\cite{Loshchilov2017DecoupledWD}, using a learning rate of $1e-6$. KL coefficient $\beta$ set to 0.04.

\mypara{Evaluation.} 
We adopt Visual CoT Benchmark~\cite{Shao2024viscot}, a comprehensive multimodal data, to measure performance across a broad range of document reasoning tasks. Following Visual CoT evaluation protocol, we also assess model’s zero-shot ability using SROIE~\cite{Huang2019ICDAR2019COsroie}, DUDE~\cite{VanLandeghem2023DocumentUDdude}, and Visual7W~\cite{Zhu2015Visual7WGQ} datasets, evaluating its capacity to generalize beyond its training distribution. We leverage the standard metrics provided by Visual CoT Benchmark~\cite{Shao2024viscot} to evaluate the model.

\subsection{Main Results}
\cref{tab:cot_benchmark} presents the results of our model \ourmodel compared to several state-of-the-art MLLMs on the Visual CoT benchmark. The evaluation spans both document-oriented understanding tasks (including text-based document comprehension and chart analysis) and general multimodal understanding (covering general VQA and relational reasoning). Our model demonstrates significant improvements over baseline models, particularly in document understanding tasks, and achieves strong generalization across multiple reasoning domains. 
Across document-oriented tasks, \ourmodel-7B (1536$^2$, 8k) achieves the highest overall scores, outperforming prior methods, including VisCoT-7B, Qwen2.5VL, and LLaVA variants. The document understanding improvements are largely attributed to the GRPO-based RL strategy with the multi-objective rewards, which enhances the model’s ability to focus on task-relevant regions within documents. In chart understanding (InfoQA), \ourmodel-7B (1536$^2$, 4k) achieves 0.689, surpassing both VisCoT-7B (336$^2$, 438k, row 5) and Qwen2.5VL-7B (1536$^2$, row 9), which score 0.324 and 0.674, respectively. This suggests that RL enables better multimodal reasoning, allowing the model to extract and interpret structured information from chart data representations.

\mypara{Zero Shot Capabilities.}
As shown in \cref{tab:cot_benchmark}, highlighted results in gray indicate zero-shot generalization performance, where training data splits were not included in the training phase. 
Our model achieves competitive zero-shot results, particularly in DUDE and SROIE requiring fine-grained text recognition and layout-aware reasoning.
Compared to Qwen2.5VL-7B (1536$^2$, row 7), which achieves 0.492 on DUDE and 0.708 on SROIE, our model reaches 0.568 on DUDE and 0.814 on SROIE. \ourmodel outperforms all non-RL-based models, demonstrating that RL with verifiable rewards substantially improves performance in text-heavy, document-based tasks.
The performance gap stems from training scale (4k vs. 438k) and zero-shot tasks compared to VisCoT. Still, DocThinker-7B ($336^2$) outperforms VisCoT-7B ($336^2$) on all non-zero-shot tasks using only 4k samples and same input.
For general multimodal understanding, including VQA and relational reasoning, \ourmodel (row 13) continues to achieve competitive performance compared to baselines on Flickr30k (0.701), Visual7W (0.625), GQA (0.694), and VSR (0.721). Particularly in relational reasoning tasks, which require understanding object interactions and contextual relationships, our RL framework enables superior performance.
The zero-shot improvements suggest that RL enhances adaptability and generalization, allowing the model to generalize beyond its training distribution and reason effectively in previously unseen scenarios.

\mypara{RL \vs SFT.}
A direct comparison between supervised fine-tuning (SFT) and RL-based models demonstrates the clear advantage of reinforcement learning. Comparing Qwen2.5VL-7B (SFT, 336$^2$, 4k, row 8) with \ourmodel-7B (RL, 336$^2$, 4k, row 12), we observe substantial gains across multiple tasks. For example, in DocQA, our model improves from 0.355 to 0.579, and in TextQA, it rises from 0.740 to 0.802. Similarly, in Visual7W, our RL model achieves 0.580 compared to 0.563 with SFT, and in GQA, it improves from 0.467 to 0.546. These improvements highlight that GRPO-based RL training enhances model reasoning and decision-making, enabling it to produce more explainable, structured, and accurate outputs. The multi-objective reward functions, particularly those focused on RoI IoU and rephrased question quality, contribute significantly to the model’s ability to focus on relevant document regions and generate clearer reasoning traces.

\begin{figure*}[htbp]
    \centering
    \centerline{\includegraphics[width=0.99\linewidth]{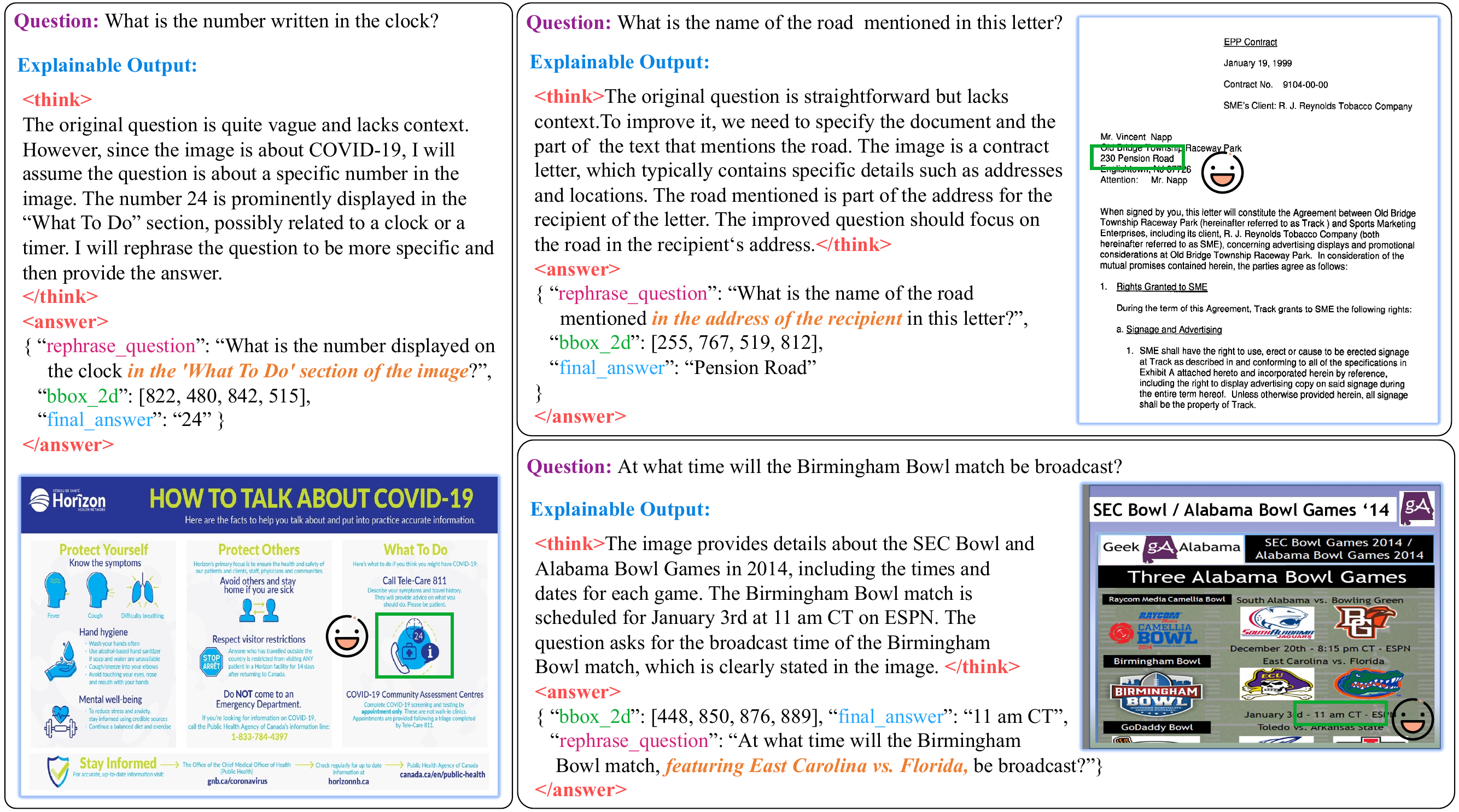}}
    \captionsetup{width=1.0\textwidth}
    \vspace{-2mm}
    \caption{\textbf{Qualitative results of \ourmodel.} The thinking process significantly improves the reasoning ability and explainability.}
    \label{fig:visualization}
    \vspace{-5mm}
\end{figure*}

\mypara{Explainability Ability.}
A key advantage of \ourmodel over existing MLLMs for document understanding is its ability to generate explicit, explainable reasoning steps, rather than merely providing direct answers. Through our reinforcement learning framework, the model systematically breaks down its thought process into structured intermediate steps, with detailed reasoning enclosed in \textless think\textgreater\textless /think\textgreater tags and the final response presented within \textless answer\textgreater\textless /answer\textgreater tags. This structured output enhances transparency, making it easier to analyze how the model arrives at its conclusions. As qualitative performance is shown in \cref{fig:visualization}, by incorporating structured reasoning rewards, including rephrased question clarity and RoI localization accuracy, our approach significantly enhances explainability over traditional SFT-based MLLMs, ensuring that responses are not only accurate but also more explainable and systematically derived.

\mypara{Visual Grounding.}
We further evaluate \ourmodel on the TextREC dataset~\cite{Gao2023TextRECAD}, which requires locating objects based on text-conditioned referring expressions. By leveraging RoI IoU reward, \ourmodel effectively aligns textual and visual information, improving grounding accuracy.
As shown in \cref{tab:visual_grounding}, \ourmodel-7B achieves 82.4\% Precision@1, surpassing specialized models like TAMN (80.8\%) and MDETR (63.3\%). The improvement highlights the impact of reinforcement learning with verifiable rewards, which enhances the model’s ability to precisely localize objects tied to scene text. These results demonstrate that \ourmodel extends beyond document understanding, excelling in spatial reasoning and multimodal grounding.

\begin{table}[ht]
    \centering
    \renewcommand\arraystretch{1}

    \setlength{\tabcolsep}{2.0mm}
     \resizebox{0.7\linewidth}{!}{
\begin{tabular}{lcc}
\toprule
Model      & Template1  & Template2 \\ 
\midrule
\multicolumn{3}{c}{Specialist Models}  \\
\midrule
TransVG~\cite{Deng2021TransVGEV} & 50.1  & 54.0 \\
MAttNet~\cite{Yu2018MAttNetMA}  & 52.3  &60.5   \\ 
QRNet~\cite{Ye2022ShiftingMAqrnet}  &    52.7 &59.1 \\ 
MDETR~\cite{Kamath2021MDETRM} & 54.4 &63.3 \\
TAMN~\cite{Gao2023TextRECAD} & 77.8 &80.8 \\
\midrule
\ourmodel-7B & \multicolumn{2}{c}{82.4} \\
\bottomrule
\end{tabular}
}
\vspace{-2mm}
\caption{\textbf{Performance on the TextREC dataset.} Precision@1 (\%)  is reported. Template1 and Template2 represent using two different ways to construct referring expressions proposed in ~\cite{Gao2023TextRECAD}. Template1 is ``The object with \textless OCR string \textgreater  ~on it'', and Template2 is ``The \textless category name \textgreater ~with \textless OCR string \textgreater ~on it''.
}
\label{tab:visual_grounding}
\vspace{-4mm}
\end{table}

\subsection{Ablation Study}

\mypara{Ablating Reward Functions.}
To assess the impact of reward functions, we ablate RoI IoU (RI) and Rephrase Question (RQ) rewards, analyzing their effect across four datasets, as provided in~\cref{tab:abla_reward_func}. Removing RoI IoU reduces performance, particularly on InfoQA and TextCaps, indicating its role in document understanding tasks requiring precise visual grounding. Eliminating Rephrase Question leads to notable drops in TextVQA and DocVQA, highlighting its importance for clarifying ambiguous queries to help the model generate more accurate and explainable responses. The most severe degradation occurs when both are removed, confirming their complementary contributions. This confirms that reinforcement learning with multi-objective rewards is crucial for enhancing both accuracy and reasoning quality in multimodal document understanding.

\begin{table}[htbp]
    \centering
    \renewcommand\arraystretch{1}
    \setlength{\tabcolsep}{2.0mm}

    \resizebox{1\linewidth}{!}{
\begin{tabular}{lcccc}

    \toprule
	\multirow{1}[0]{*}{Method} & 

	\multicolumn{1}{c}{ DocVQA } &
	\multicolumn{1}{c}{ TextCaps} &
	\multicolumn{1}{c}{TextVQA} &
	\multicolumn{1}{c}{InfoQA} 
	 \\

    \midrule
	\ourmodel-7B  & 0.795 &  0.738  & 0.827 & 0.689  \\
    \midrule
	w/o RoI IoU   & 0.775 & 0.693  & 0.803  & 0.637  \\
w/o Rephrase Question  & 0.763 & 0.716  & 0.772  & 0.658  \\
	w/o RI \& RQ  & 0.741 & 0.662  & 0.758  & 0.602  \\
    \bottomrule
\end{tabular}
}
\vspace{-2mm}
    \caption{
    \textbf{Ablation study of reward functions.} The \ourmodel uses 1536$^2$ input size with 4data4k training data setting. RI and RQ short for RoI IoU and rephrase question reward, respectively.
    }
    \label{tab:abla_reward_func}
    \vspace{-3mm}
\end{table}

\mypara{Ablating KL Divergence.}
To evaluate the impact of KL divergence regularization, we analyze different values of the coefficient \(\beta\) and its effect on performance across four document-oriented datasets. 
As shown in ~\cref{tab:abla_kl_coef}, when KL regularization is entirely removed (\(\beta = 0\)), the model exhibits a decline in performance across all datasets, particularly in TextVQA and DocVQA, indicating that KL regularization plays a key role in stabilizing training and preventing catastrophic forgetting during reinforcement learning. Introducing a small KL weight (\(\beta = 0.001\)) improves performance slightly but does not fully match the stability compared to setting $\beta = 0.04$. The results suggest that an appropriate KL divergence coefficient is crucial for maintaining balance between exploration and stability, ensuring robust performance across datasets.

\begin{table}[htbp]
    \centering
    \renewcommand\arraystretch{1}
    \setlength{\tabcolsep}{2.0mm}

    \resizebox{1\linewidth}{!}{
\begin{tabular}{lcccc}

    \toprule
	\multirow{1}[0]{*}{Method} & 

	\multicolumn{1}{c}{ DocVQA } &
	\multicolumn{1}{c}{ TextCaps} &
	\multicolumn{1}{c}{TextVQA} &
	\multicolumn{1}{c}{InfoQA} 
	 \\

    \midrule
	\ourmodel-7B ($\beta=0.04$) & 0.795 &  0.738  & 0.827 & 0.689  \\
    \midrule
	w/o KL ($\beta=0$)          & 0.780 & 0.719  & 0.803  & 0.676  \\
      $\beta=0.001 $                 & 0.785 & 0.726  & 0.812  & 0.682  \\
    \bottomrule
\end{tabular}
}
\vspace{-2mm}
    \caption{
    \textbf{Ablation study of the effect of KL Divergence.} The \ourmodel uses 1536$^2$ input size with 4data4k training data setting.
    }
    \label{tab:abla_kl_coef}
    \vspace{-6mm}
\end{table}

\section{Conclusions and Future Works}
In this work, we introduced \ourmodel, a reinforcement learning-based framework designed to enhance explainability, adaptability, and reasoning ability in multimodal document understanding. By leveraging Group Relative Policy Optimization and a multi-objective reward system, our approach dynamically refines reasoning strategies at inference time, overcoming the limitations of static Chain-of-Thought reasoning and supervised fine-tuning. \ourmodel achieves state-of-the-art or highly competitive performance on standard benchmarks compared with previous SFT-based methods across multiple document understanding tasks. Expanding \ourmodel to larger multimodal models and broader reasoning tasks and scenarios, such as scientific and legal document analysis, could further improve adaptability.

\noindent \textbf{Acknowledgements.} This work was supported by the NSFC (62225603, 62206104), and Alibaba AIR program.

{
    \small
    \bibliographystyle{ieeenat_fullname}
    \bibliography{main}

\begin{thebibliography}{22}
\providecommand{\natexlab}[1]{#1}
\providecommand{\url}[1]{\texttt{#1}}
\expandafter\ifx\csname urlstyle\endcsname\relax
  \providecommand{\doi}[1]{doi: #1}\else
  \providecommand{\doi}{doi: \begingroup \urlstyle{rm}\Url}\fi

\bibitem[Chen et~al.(2024{\natexlab{a}})Chen, Wang, Tian, Ye, Gao, Cui, Tong, Hu, Luo, Ma, Ma, Wang, wen Dong, Yan, Guo, He, Jin, Xu, Wang, Wei, Li, Zhang, Zhang, Lu, Zhu, Lu, Lin, and Qiao]{Chen2024HowFAinternvl2}
Zhe Chen, Weiyun Wang, Hao Tian, Shenglong Ye, Zhangwei Gao, Erfei Cui, Wenwen Tong, Kongzhi Hu, Jiapeng Luo, Zheng Ma, Ji Ma, Jiaqi Wang, Xiao wen Dong, Hang Yan, Hewei Guo, Conghui He, Zhenjiang Jin, Chaochao Xu, Bin Wang, Xingjian Wei, Wei Li, Wenjian Zhang, Bo Zhang, Lewei Lu, Xizhou Zhu, Tong Lu, Dahua Lin, and Yu Qiao.
\newblock How far are we to gpt-4v? closing the gap to commercial multimodal models with open-source suites.
\newblock \emph{Science China Information Sciences}, 2024{\natexlab{a}}.

\bibitem[Chen et~al.(2024{\natexlab{b}})Chen, Wu, Wang, Su, Chen, Xing, Muyan, Zhang, Zhu, Lu, Li, Luo, Lu, Qiao, and Dai]{Chen2023InternVLS}
Zhe Chen, Jiannan Wu, Wenhai Wang, Weijie Su, Guo Chen, Sen Xing, Zhong Muyan, Qinglong Zhang, Xizhou Zhu, Lewei Lu, Bin Li, Ping Luo, Tong Lu, Yu Qiao, and Jifeng Dai.
\newblock Intern vl: Scaling up vision foundation models and aligning for generic visual-linguistic tasks.
\newblock In \emph{IEEE/CVF Conference on Computer Vision and Pattern Recognition}, pages 24185--24198, 2024{\natexlab{b}}.

\bibitem[Dai et~al.(2023)Dai, Li, Li, Tiong, Zhao, Wang, Li, Fung, and Hoi]{Dai2023InstructBLIPTG}
Wenliang Dai, Junnan Li, Dongxu Li, Anthony Meng~Huat Tiong, Junqi Zhao, Weisheng Wang, Boyang~Albert Li, Pascale Fung, and Steven C.~H. Hoi.
\newblock Instructblip: Towards general-purpose vision-language models with instruction tuning.
\newblock In \emph{Neural Information Processing Systems}, 2023.

\bibitem[Feng et~al.(2023)Feng, Wang, Tang, Lu, gang Zhou, Li, and Huang]{Feng2023UniDocAU}
Hao Feng, Zijian Wang, Jingqun Tang, Jinghui Lu, Wen gang Zhou, Houqiang Li, and Can Huang.
\newblock Unidoc: A universal large multimodal model for simultaneous text detection, recognition, spotting and understanding.
\newblock \emph{ArXiv}, abs/2308.11592, 2023.

\bibitem[Feng et~al.(2024)Feng, Liu, Liu, Tang, Zhou, Li, and Huang]{feng2024docpedia}
Hao Feng, Qi Liu, Hao Liu, Jingqun Tang, Wengang Zhou, Houqiang Li, and Can Huang.
\newblock Docpedia: Unleashing the power of large multimodal model in the frequency domain for versatile document understanding.
\newblock \emph{Science China Information Sciences}, 67\penalty0 (12):\penalty0 1--14, 2024.

\bibitem[Hu et~al.(2024)Hu, Xu, Li, Li, Chen, and Tu]{Hu2023BLIVAAS}
Wenbo Hu, Y. Xu, Y. Li, W. Li, Z. Chen, and Z. Tu.
\newblock Bliva: A simple multimodal llm for better handling of text-rich visual questions.
\newblock In \emph{Proceedings of the AAAI Conference on Artificial Intelligence}, 2024.

\bibitem[Huang et~al.(2025)Huang, Liu, Liang, Jin, and Bai]{Huang2024MiniMonkeyMA}
Mingxin Huang, Yuliang Liu, Dingkang Liang, Lianwen Jin, and Xiang Bai.
\newblock Mini-monkey: Multi-scale adaptive cropping for multimodal large language models.
\newblock In \emph{ICLR}, pages 1--15, 2025.

\bibitem[Li et~al.(2023)Li, Li, Savarese, and Hoi]{Li2023BLIP2BL}
Junnan Li, Dongxu Li, Silvio Savarese, and Steven C.~H. Hoi.
\newblock Blip-2: Bootstrapping language-image pre-training with frozen image encoders and large language models.
\newblock In \emph{International Conference on Machine Learning}, 2023.

\bibitem[Li et~al.(2024)Li, Yang, Liu, Ma, Zhang, Yang, Sun, Liu, and Bai]{Li2023MonkeyIR}
Zhang Li, Biao Yang, Qiang Liu, Zhiyin Ma, Shuo Zhang, Jingxu Yang, Yabo Sun, Yuliang Liu, and Xiang Bai.
\newblock Monkey: Image resolution and text label are important things for large multi-modal models.
\newblock In \emph{IEEE/CVF Conference on Computer Vision and Pattern Recognition}, pages 26753--26763, 2024.

\bibitem[Liu et~al.(2024{\natexlab{a}})Liu, Guan, Wu, Li, Chen, Yacoob, Manocha, and Zhou]{Liu2023HallusionBenchYS}
Fuxiao Liu, Tianrui Guan, Xiyang Wu, Zongxia Li, Lichang Chen, Yaser Yacoob, Dinesh Manocha, and Tianyi Zhou.
\newblock Hallusionbench: You see what you think? or you think what you see? an image-context reasoning benchmark challenging for gpt-4v(ision), llava-1.5, and other multi-modality models.
\newblock In \emph{IEEE/CVF Conference on Computer Vision and Pattern Recognition}, 2024{\natexlab{a}}.

\bibitem[Liu et~al.(2024{\natexlab{b}})Liu, Li, Li, Li, Zhang, Shen, and Lee]{liu2024llavanextllava1.5}
Haotian Liu, Chunyuan Li, Yuheng Li, Bo Li, Yuanhan Zhang, Sheng Shen, and Yong~Jae Lee.
\newblock Llava-next: Improved reasoning, ocr, and world knowledge, 2024{\natexlab{b}}.

\bibitem[Liu et~al.(2024{\natexlab{c}})Liu, Li, Huang, Yang, Yu, Li, Yin, lin Liu, Jin, and Bai]{Liu2023OCRBenchOT}
Yuliang Liu, Zhang Li, Mingxin Huang, Biao Yang, Wenwen Yu, Chunyuan Li, Xucheng Yin, Cheng lin Liu, Lianwen Jin, and Xiang Bai.
\newblock Ocrbench: on the hidden mystery of ocr in large multimodal models.
\newblock \emph{Science China Information Sciences}, 2024{\natexlab{c}}.

\bibitem[Liu et~al.(2024{\natexlab{d}})Liu, Yang, Liu, Li, Ma, Zhang, and Bai]{Liu2024TextMonkeyAO}
Yuliang Liu, Biao Yang, Qiang Liu, Zhang Li, Zhiyin Ma, Shuo Zhang, and Xiang Bai.
\newblock Textmonkey: An ocr-free large multimodal model for understanding document.
\newblock \emph{ArXiv}, abs/2403.04473, 2024{\natexlab{d}}.

\bibitem[Schulman(2020)]{Schulmankl}
John Schulman.
\newblock {Approximating KL Divergence}.
\newblock \url{http://joschu.net/blog/kl-approx.html}, 2020.

\bibitem[Schulman et~al.(2017)Schulman, Wolski, Dhariwal, Radford, and Klimov]{PPO}
John Schulman, Filip Wolski, Prafulla Dhariwal, Alec Radford, and Oleg Klimov.
\newblock Proximal policy optimization algorithms.
\newblock \emph{arXiv:1707.06347}, 2017.

\bibitem[Shao et~al.(2024{\natexlab{a}})Shao, Qian, Xiao, Song, Zong, Wang, Liu, and Li]{Shao2024viscot}
Hao Shao, Shengju Qian, Han Xiao, Guanglu Song, Zhuofan Zong, Letian Wang, Yu Liu, and Hongsheng Li.
\newblock Visual cot: Advancing multi-modal language models with a comprehensive dataset and benchmark for chain-of-thought reasoning.
\newblock In \emph{Neural Information Processing Systems}, 2024{\natexlab{a}}.

\bibitem[Shao et~al.(2024{\natexlab{b}})Shao, Wang, Zhu, Xu, Song, Zhang, Li, Wu, and Guo]{Shao2024DeepSeekMathPT}
Zhihong Shao, Peiyi Wang, Qihao Zhu, Runxin Xu, Jun-Mei Song, Mingchuan Zhang, Y.~K. Li, Yu Wu, and Daya Guo.
\newblock Deepseekmath: Pushing the limits of mathematical reasoning in open language models.
\newblock \emph{ArXiv}, abs/2402.03300, 2024{\natexlab{b}}.

\bibitem[Wang et~al.(2023)Wang, gang Zhou, Feng, Zhou, and Li]{Wang2023TowardsIDtgdoc}
Yonghui Wang, Wen gang Zhou, Hao Feng, Keyi Zhou, and Houqiang Li.
\newblock Towards improving document understanding: An exploration on text-grounding via mllms.
\newblock \emph{ArXiv}, abs/2311.13194, 2023.

\bibitem[wen Dong et~al.(2024)wen Dong, Zhang, Zang, Cao, Wang, Ouyang, Wei, Zhang, Duan, Cao, Zhang, Li, Yan, Gao, Zhang, Li, Li, Chen, He, Zhang, Qiao, Lin, and Wang]{Dong2024InternLMXComposer2MF}
Xiao wen Dong, Pan Zhang, Yuhang Zang, Yuhang Cao, Bin Wang, Linke Ouyang, Xilin Wei, Songyang Zhang, Haodong Duan, Maosong Cao, Wenwei Zhang, Yining Li, Hang Yan, Yang Gao, Xinyue Zhang, Wei Li, Jingwen Li, Kai Chen, Conghui He, Xingcheng Zhang, Yu Qiao, Dahua Lin, and Jiaqi Wang.
\newblock Internlm-xcomposer2: Mastering free-form text-image composition and comprehension in vision-language large model.
\newblock \emph{ArXiv}, abs/2401.16420, 2024.

\bibitem[Ye et~al.(2023)Ye, Xu, Xu, Ye, Yan, Zhou, Wang, Hu, Shi, Shi, Li, Xu, Chen, Tian, Qi, Zhang, and Huang]{Ye2023mPLUGOwlME}
Qinghao Ye, Haiyang Xu, Guohai Xu, Jiabo Ye, Ming Yan, Yi Zhou, Junyan Wang, Anwen Hu, Pengcheng Shi, Yaya Shi, Chenliang Li, Yuanhong Xu, Hehong Chen, Junfeng Tian, Qiang Qi, Ji Zhang, and Feiyan Huang.
\newblock mplug-owl: Modularization empowers large language models with multimodality.
\newblock \emph{ArXiv}, abs/2304.14178, 2023.

\bibitem[Ye et~al.(2024)Ye, Xu, Ye, Yan, Hu, Liu, Qian, Zhang, Huang, and Zhou]{Ye2023mPLUGOwI2RM}
Qinghao Ye, Haiyang Xu, Jiabo Ye, Mingshi Yan, Anwen Hu, Haowei Liu, Qi Qian, Ji Zhang, Fei Huang, and Jingren Zhou.
\newblock mplug-owi2: Revolutionizing multi-modal large language model with modality collaboration.
\newblock In \emph{IEEE/CVF Conference on Computer Vision and Pattern Recognition}, pages 13040--13051, 2024.

\bibitem[Zhang et~al.(2023)Zhang, Zhang, Gu, Zhou, Lipka, Yang, and Sun]{zhang2023llavar}
Yanzhe Zhang, Ruiyi Zhang, Jiuxiang Gu, Yufan Zhou, Nedim Lipka, Diyi Yang, and Tong Sun.
\newblock Llavar: Enhanced visual instruction tuning for text-rich image understanding.
\newblock \emph{arXiv preprint arXiv:2306.17107}, 2023.

\end{thebibliography}


\begin{thebibliography}{54}
\providecommand{\natexlab}[1]{#1}
\providecommand{\url}[1]{\texttt{#1}}
\expandafter\ifx\csname urlstyle\endcsname\relax
  \providecommand{\doi}[1]{doi: #1}\else
  \providecommand{\doi}{doi: \begingroup \urlstyle{rm}\Url}\fi

\bibitem[Bai et~al.(2025)Bai, Chen, Liu, Wang, Ge, Song, Dang, Wang, Wang, Tang, Zhong, Zhu, Yang, Li, Wan, Wang, Ding, Fu, Xu, Ye, Zhang, Xie, Cheng, Zhang, Yang, Xu, and Lin]{Shuai2025Qwen2.5-VL}
Shuai Bai, Keqin Chen, Xuejing Liu, Jialin Wang, Wenbin Ge, Sibo Song, Kai Dang, Peng Wang, Shijie Wang, Jun Tang, Humen Zhong, Yuanzhi Zhu, Mingkun Yang, Zhaohai Li, Jianqiang Wan, Pengfei Wang, Wei Ding, Zheren Fu, Yiheng Xu, Jiabo Ye, Xi Zhang, Tianbao Xie, Zesen Cheng, Hang Zhang, Zhibo Yang, Haiyang Xu, and Junyang Lin.
\newblock Qwen2.5-vl technical report.
\newblock \emph{arXiv preprint arXiv:2502.13923}, 2025.

\bibitem[Chen et~al.(2024)Chen, Wang, Tian, Ye, Gao, Cui, Tong, Hu, Luo, Ma, et~al.]{chen2024far}
Zhe Chen, Weiyun Wang, Hao Tian, Shenglong Ye, Zhangwei Gao, Erfei Cui, Wenwen Tong, Kongzhi Hu, Jiapeng Luo, Zheng Ma, et~al.
\newblock How far are we to gpt-4v? closing the gap to commercial multimodal models with open-source suites.
\newblock \emph{Science China Information Sciences}, 67\penalty0 (12):\penalty0 220101, 2024.

\bibitem[Chu et~al.(2025)Chu, Zhai, Yang, Tong, Xie, Schuurmans, Le, Levine, and Ma]{Chu2025SFTMR}
Tianzhe Chu, Yuexiang Zhai, Jihan Yang, Shengbang Tong, Saining Xie, Dale Schuurmans, Quoc~V. Le, Sergey Levine, and Yi Ma.
\newblock Sft memorizes, rl generalizes: A comparative study of foundation model post-training.
\newblock \emph{ArXiv}, abs/2501.17161, 2025.

\bibitem[Dang et~al.(2024)Dang, Huang, Huo, Yan, Huang, Liu, Gao, Zhang, Qian, Wang, et~al.]{dang2024explainable}
Yunkai Dang, Kaichen Huang, Jiahao Huo, Yibo Yan, Sirui Huang, Dongrui Liu, Mengxi Gao, Jie Zhang, Chen Qian, Kun Wang, et~al.
\newblock Explainable and interpretable multimodal large language models: A comprehensive survey.
\newblock \emph{ArXiv}, arXiv:2412.02104, 2024.

\bibitem[DeepSeek-AI et~al.(2025)DeepSeek-AI, Guo, Yang, Zhang, Song, Zhang, Xu, Zhu, Ma, Wang, Bi, Zhang, Yu, Wu, Wu, Gou, Shao, Li, Gao, Liu, Xue, Wang, Wu, Feng, Lu, Zhao, Deng, Zhang, Ruan, Dai, Chen, Ji, Li, Lin, Dai, Luo, Hao, Chen, Li, Zhang, Bao, Xu, Wang, Ding, Xin, Gao, Qu, Li, Guo, Li, Wang, Chen, Yuan, Qiu, Li, Cai, Ni, Liang, Chen, Dong, Hu, Gao, Guan, Huang, Yu, Wang, Zhang, Zhao, Wang, Zhang, Xu, Xia, Zhang, Zhang, Tang, Li, Wang, Li, Tian, Huang, Zhang, Wang, Chen, Du, Ge, Zhang, Pan, Wang, Chen, Jin, Chen, Lu, Zhou, Chen, Ye, Wang, Yu, Zhou, Pan, Li, Zhou, Wu, Yun, Pei, Sun, Wang, Zeng, Zhao, Liu, Liang, Gao, Yu, Zhang, Xiao, An, Liu, Wang, Chen, Nie, Cheng, Liu, Xie, Liu, Yang, Li, Su, Lin, Li, Jin, Shen, Chen, Sun, Wang, Song, Zhou, Wang, Shan, Li, Wang, Wei, Zhang, Xu, Li, Zhao, Sun, Wang, Yu, Zhang, Shi, Xiong, He, Piao, Wang, Tan, Ma, Liu, Guo, Ou, Wang, Gong, Zou, He, Xiong, Luo, mei You, Liu, Zhou, Zhu, Huang, Li, Zheng, Zhu, Ma, Tang, Zha, Yan, Ren, Ren, Sha, Fu, Xu, Xie, guo Zhang,
  Hao, Ma, Yan, Wu, Gu, Zhu, Liu, Li, Xie, Song, Pan, Huang, Xu, Zhang, and Zhang]{DeepSeekAI2025DeepSeekR1IR}
DeepSeek-AI, Daya Guo, Dejian Yang, Haowei Zhang, Jun-Mei Song, Ruoyu Zhang, Runxin Xu, Qihao Zhu, Shirong Ma, Peiyi Wang, Xiaoling Bi, Xiaokang Zhang, Xingkai Yu, Yu Wu, Z.~F. Wu, Zhibin Gou, Zhihong Shao, Zhuoshu Li, Ziyi Gao, Aixin Liu, Bing Xue, Bing-Li Wang, Bochao Wu, Bei Feng, Chengda Lu, Chenggang Zhao, Chengqi Deng, Chenyu Zhang, Chong Ruan, Damai Dai, Deli Chen, Dong-Li Ji, Erhang Li, Fangyun Lin, Fucong Dai, Fuli Luo, Guangbo Hao, Guanting Chen, Guowei Li, H. Zhang, Han Bao, Hanwei Xu, Haocheng Wang, Honghui Ding, Huajian Xin, Huazuo Gao, Hui Qu, Hui Li, Jianzhong Guo, Jiashi Li, Jiawei Wang, Jingchang Chen, Jingyang Yuan, Junjie Qiu, Junlong Li, Jiong Cai, Jiaqi Ni, Jian Liang, Jin Chen, Kai Dong, Kai Hu, Kaige Gao, Kang Guan, Kexin Huang, Kuai Yu, Lean Wang, Lecong Zhang, Liang Zhao, Litong Wang, Liyue Zhang, Lei Xu, Leyi Xia, Mingchuan Zhang, Minghua Zhang, M. Tang, Meng Li, Miaojun Wang, Mingming Li, Ning Tian, Panpan Huang, Peng Zhang, Qiancheng Wang, Qinyu Chen, Qiushi Du, Ruiqi Ge, Ruisong
  Zhang, Ruizhe Pan, Runji Wang, R.~J. Chen, R.~L. Jin, Ruyi Chen, Shanghao Lu, Shangyan Zhou, Shanhuang Chen, Shengfeng Ye, Shiyu Wang, Shuiping Yu, Shunfeng Zhou, Shuting Pan, S.~S. Li, Shuang Zhou, Shao-Kang Wu, Tao Yun, Tian Pei, Tianyu Sun, T. Wang, Wangding Zeng, Wanjia Zhao, Wen Liu, Wenfeng Liang, Wenjun Gao, Wen-Xia Yu, Wentao Zhang, W.~L. Xiao, Wei An, Xiaodong Liu, Xiaohan Wang, Xiaokang Chen, Xiaotao Nie, Xin Cheng, Xin Liu, Xin Xie, Xingchao Liu, Xinyu Yang, Xinyuan Li, Xuecheng Su, Xuheng Lin, X.~Q. Li, Xiangyu Jin, Xi-Cheng Shen, Xiaosha Chen, Xiaowen Sun, Xiaoxiang Wang, Xinnan Song, Xinyi Zhou, Xianzu Wang, Xinxia Shan, Y.~K. Li, Y.~Q. Wang, Y.~X. Wei, Yang Zhang, Yanhong Xu, Yao Li, Yao Zhao, Yaofeng Sun, Yaohui Wang, Yi Yu, Yichao Zhang, Yifan Shi, Yi Xiong, Ying He, Yishi Piao, Yisong Wang, Yixuan Tan, Yiyang Ma, Yiyuan Liu, Yongqiang Guo, Yuan Ou, Yuduan Wang, Yue Gong, Yu-Jing Zou, Yujia He, Yunfan Xiong, Yu-Wei Luo, Yu mei You, Yuxuan Liu, Yuyang Zhou, Y.~X. Zhu, Yanping Huang, Yao Li,
  Yi Zheng, Yuchen Zhu, Yunxiang Ma, Ying Tang, Yukun Zha, Yuting Yan, Zehui Ren, Zehui Ren, Zhangli Sha, Zhe Fu, Zhean Xu, Zhenda Xie, Zhen guo Zhang, Zhewen Hao, Zhicheng Ma, Zhigang Yan, Zhiyu Wu, Zihui Gu, Zijia Zhu, Zijun Liu, Zi-An Li, Ziwei Xie, Ziyang Song, Zizheng Pan, Zhen Huang, Zhipeng Xu, Zhongyu Zhang, and Zhen Zhang.
\newblock Deepseek-r1: Incentivizing reasoning capability in llms via reinforcement learning.
\newblock \emph{ArXiv}, abs/2501.12948, 2025.

\bibitem[Deng et~al.(2021)Deng, Yang, Chen, gang Zhou, and Li]{Deng2021TransVGEV}
Jiajun Deng, Zhengyuan Yang, Tianlang Chen, Wen gang Zhou, and Houqiang Li.
\newblock Transvg: End-to-end visual grounding with transformers.
\newblock In \emph{IEEE/CVF International Conference on Computer Vision}, pages 1749--1759, 2021.

\bibitem[Feng et~al.(2024)Feng, Liu, Liu, Tang, Zhou, Li, and Huang]{feng2024docpedia}
Hao Feng, Qi Liu, Hao Liu, Jingqun Tang, Wengang Zhou, Houqiang Li, and Can Huang.
\newblock Docpedia: Unleashing the power of large multimodal model in the frequency domain for versatile document understanding.
\newblock \emph{Science China Information Sciences}, 67\penalty0 (12):\penalty0 1--14, 2024.

\bibitem[Fu et~al.(2025)Fu, Liu, Yang, Corring, Lu, Yang, Roth, Flor{\^e}ncio, and Zhang]{Fu2025ReFocusVE}
Xingyu Fu, Minqian Liu, Zhengyuan Yang, John Corring, Yijuan Lu, Jianwei Yang, Dan Roth, Dinei A.~F. Flor{\^e}ncio, and Cha Zhang.
\newblock Refocus: Visual editing as a chain of thought for structured image understanding.
\newblock \emph{ArXiv}, abs/2501.05452, 2025.

\bibitem[Gao et~al.(2023)Gao, Yang, Wang, Yang, Yu, Liu, and Bai]{Gao2023TextRECAD}
Chenyu Gao, Biao Yang, Hao Wang, Mingkun Yang, Wenwen Yu, Yuliang Liu, and Xiang Bai.
\newblock Textrec: A dataset for referring expression comprehension with reading comprehension.
\newblock In \emph{IEEE International Conference on Document Analysis and Recognition}, 2023.

\bibitem[Hsieh et~al.(2024)Hsieh, Bi, Jiang, Liu, Peng, Zhang, Pan, Xu, Wang, Chen, Feng, Wen, Song, Wang, Liu, Yang, Li, Jing, Ren, Song, Tseng, Zhang, Yan, Niu, Chen, Wang, and Liang]{Hsieh2024ACG}
Weiche Hsieh, Ziqian Bi, Chuanqi Jiang, Junyu Liu, Benji Peng, Sen Zhang, Xuanhe Pan, Jiawei Xu, Jinlang Wang, Keyu Chen, Pohsun Feng, Yizhu Wen, Xinyuan Song, Tianyang Wang, Ming Liu, Junjie Yang, Ming Li, Bowen Jing, Jintao Ren, Jun-Jie Song, Hong-Ming Tseng, Yichao Zhang, Lawrence~K.Q. Yan, Qian Niu, Silin Chen, Yunze Wang, and Chia~Xin Liang.
\newblock A comprehensive guide to explainable ai: From classical models to llms.
\newblock \emph{ArXiv}, abs/2412.00800, 2024.

\bibitem[Huang et~al.(2019)Huang, Chen, He, Bai, Karatzas, Lu, and Jawahar]{Huang2019ICDAR2019COsroie}
Zheng Huang, Kai Chen, Jianhua He, Xiang Bai, Dimosthenis Karatzas, Shijian Lu, and C.~V. Jawahar.
\newblock Icdar2019 competition on scanned receipt ocr and information extraction.
\newblock In \emph{International Conference on Document Analysis and Recognition}, pages 1516--1520, 2019.

\bibitem[Hudson and Manning(2019)]{Hudson2019GQAANgqa}
Drew~A. Hudson and Christopher~D. Manning.
\newblock Gqa: A new dataset for real-world visual reasoning and compositional question answering.
\newblock In \emph{IEEE/CVF Conference on Computer Vision and Pattern Recognition}, pages 6693--6702, 2019.

\bibitem[Jaech et~al.(2024)Jaech, Kalai, Lerer, Richardson, El-Kishky, Low, Helyar, Madry, Beutel, Carney, et~al.]{jaech2024openai}
Aaron Jaech, Adam Kalai, Adam Lerer, Adam Richardson, Ahmed El-Kishky, Aiden Low, Alec Helyar, Aleksander Madry, Alex Beutel, Alex Carney, et~al.
\newblock Openai o1 system card.
\newblock \emph{arXiv preprint arXiv:2412.16720}, 2024.

\bibitem[Kamath et~al.(2021)Kamath, Singh, LeCun, Misra, Synnaeve, and Carion]{Kamath2021MDETRM}
Aishwarya Kamath, Mannat Singh, Yann LeCun, Ishan Misra, Gabriel Synnaeve, and Nicolas Carion.
\newblock Mdetr - modulated detection for end-to-end multi-modal understanding.
\newblock In \emph{IEEE/CVF International Conference on Computer Vision}, pages 1760--1770, 2021.

\bibitem[Kuznetsova et~al.(2018)Kuznetsova, Rom, Alldrin, Uijlings, Krasin, Pont-Tuset, Kamali, Popov, Malloci, Kolesnikov, Duerig, and Ferrari]{Kuznetsova2018TheOIopenimage}
Alina Kuznetsova, Hassan Rom, Neil~Gordon Alldrin, Jasper R.~R. Uijlings, Ivan Krasin, Jordi Pont-Tuset, Shahab Kamali, Stefan Popov, Matteo Malloci, Alexander Kolesnikov, Tom Duerig, and Vittorio Ferrari.
\newblock The open images dataset v4.
\newblock \emph{International Journal of Computer Vision}, 128:\penalty0 1956 -- 1981, 2018.

\bibitem[Landeghem et~al.(2023)Landeghem, Tito, Łukasz Borchmann, Pietruszka, J'oziak, Powalski, Jurkiewicz, Coustaty, Ackaert, Valveny, Blaschko, Moens, and Stanislawek]{VanLandeghem2023DocumentUDdude}
Jordy~Van Landeghem, Rub{\`e}n~P{\'e}rez Tito, Łukasz Borchmann, Michal Pietruszka, Pawel J'oziak, Rafal Powalski, Dawid Jurkiewicz, Micka{\"e}l Coustaty, Bertrand Ackaert, Ernest Valveny, Matthew~B. Blaschko, Sien Moens, and Tomasz Stanislawek.
\newblock Document understanding dataset and evaluation (dude).
\newblock In \emph{IEEE/CVF International Conference on Computer Vision}, pages 19471--19483, 2023.

\bibitem[Li et~al.(2025)Li, Wu, Zhang, Xia, Mao, Dong, Vuli'c, and Wei]{Li2025mvot}
Chengzu Li, Wenshan Wu, Huanyu Zhang, Yan Xia, Shaoguang Mao, Li Dong, Ivan Vuli'c, and Furu Wei.
\newblock Imagine while reasoning in space: Multimodal visualization-of-thought.
\newblock \emph{ArXiv}, abs/2501.07542, 2025.

\bibitem[Lin et~al.(2023)Lin, Liu, Zhang, Gao, Qiu, Xiao, Qiu, Lin, Shao, Chen, Han, Huang, Zhang, He, Li, and Qiao]{lin2023sphinx}
Ziyi Lin, Chris Liu, Renrui Zhang, Peng Gao, Longtian Qiu, Han Xiao, Han Qiu, Chen Lin, Wenqi Shao, Keqin Chen, Jiaming Han, Siyuan Huang, Yichi Zhang, Xuming He, Hongsheng Li, and Yu~Jiao Qiao.
\newblock Sphinx: The joint mixing of weights, tasks, and visual embeddings for multi-modal large language models.
\newblock \emph{ArXiv}, abs/2311.07575, 2023.

\bibitem[Liu et~al.(2024{\natexlab{a}})Liu, Wei, Chen, Kong, Ge, Zhu, Zhao, Sun, Han, and Zhang]{liu2024focus}
Chenglong Liu, Haoran Wei, Jinyue Chen, Lingyu Kong, Zheng Ge, Zining Zhu, Liang Zhao, Jianjian Sun, Chunrui Han, and Xiangyu Zhang.
\newblock Focus anywhere for fine-grained multi-page document understanding.
\newblock \emph{arXiv preprint arXiv:2405.14295}, 2024{\natexlab{a}}.

\bibitem[Liu et~al.(2024{\natexlab{b}})Liu, Zeng, Liu, Yan, He, Wang, Yan, Liu, and Zhou]{liu2024skywork}
Chris~Yuhao Liu, Liang Zeng, Jiacai Liu, Rui Yan, Jujie He, Chaojie Wang, Shuicheng Yan, Yang Liu, and Yahui Zhou.
\newblock {Skywork-Reward}: Bag of tricks for reward modeling in llms.
\newblock \emph{arXiv preprint arXiv:2410.18451}, 2024{\natexlab{b}}.

\bibitem[Liu et~al.(2022)Liu, Emerson, and Collier]{Liu2022VisualSRvsr}
Fangyu Liu, Guy Edward~Toh Emerson, and Nigel Collier.
\newblock Visual spatial reasoning.
\newblock \emph{Transactions of the Association for Computational Linguistics}, 11:\penalty0 635--651, 2022.

\bibitem[Liu et~al.(2023)Liu, Li, Li, and Lee]{liu2023improvedllava}
Haotian Liu, Chunyuan Li, Yuheng Li, and Yong~Jae Lee.
\newblock Improved baselines with visual instruction tuning.
\newblock \emph{IEEE/CVF Conference on Computer Vision and Pattern Recognition}, pages 26286--26296, 2023.

\bibitem[Liu et~al.(2024{\natexlab{c}})Liu, Yang, Liu, Li, Ma, Zhang, and Bai]{Liu2024TextMonkeyAO}
Yuliang Liu, Biao Yang, Qiang Liu, Zhang Li, Zhiyin Ma, Shuo Zhang, and Xiang Bai.
\newblock Textmonkey: An ocr-free large multimodal model for understanding document.
\newblock \emph{ArXiv}, abs/2403.04473, 2024{\natexlab{c}}.

\bibitem[Liu et~al.(2025)Liu, Sun, Zang, Dong, Cao, Duan, Lin, and Wang]{Liu2025VisualRFTVR}
Ziyu Liu, Zeyi Sun, Yuhang Zang, Xiaoyi Dong, Yuhang Cao, Haodong Duan, Dahua Lin, and Jiaqi Wang.
\newblock Visual-rft: Visual reinforcement fine-tuning.
\newblock \emph{ArXiv}, 2025.

\bibitem[Loshchilov and Hutter(2017)]{Loshchilov2017DecoupledWD}
Ilya Loshchilov and Frank Hutter.
\newblock Decoupled weight decay regularization.
\newblock In \emph{International Conference on Learning Representations}, 2017.

\bibitem[Lyu et~al.(2024)Lyu, Li, Zhou, Ma, Wan, Xie, Wu, Zhang, Yao, Ding, and Wang]{Lyu2024StrucTexTv3AE}
Pengyuan Lyu, Yulin Li, Hao Zhou, Weihong Ma, Xingyu Wan, Qunyi Xie, Liang Wu, Chengquan Zhang, Kun Yao, Errui Ding, and Jingdong Wang.
\newblock Structextv3: An efficient vision-language model for text-rich image perception, comprehension, and beyond.
\newblock \emph{ArXiv}, abs/2405.21013, 2024.

\bibitem[Mathew et~al.(2020)Mathew, Karatzas, Manmatha, and Jawahar]{Mathew2020DocVQAAD}
Minesh Mathew, Dimosthenis Karatzas, R. Manmatha, and C.~V. Jawahar.
\newblock Docvqa: A dataset for vqa on document images.
\newblock In \emph{IEEE Winter Conference on Applications of Computer Vision}, pages 2199--2208, 2020.

\bibitem[Mathew et~al.(2021)Mathew, Bagal, Tito, Karatzas, Valveny, and Jawahar]{Mathew2021InfographicVQA}
Minesh Mathew, Viraj Bagal, Rub{\`e}n~P{\'e}rez Tito, Dimosthenis Karatzas, Ernest Valveny, and C.V. Jawahar.
\newblock Infographicvqa.
\newblock In \emph{IEEE/CVF Winter Conference on Applications of Computer Vision}, pages 2582--2591, 2021.

\bibitem[OpenAI(2023{\natexlab{a}})]{ChatGPT}
OpenAI.
\newblock {ChatGPT}.
\newblock \url{https://openai.com/chatgpt}, 2023{\natexlab{a}}.
\newblock Accessed: 2025-02-20.

\bibitem[OpenAI(2023{\natexlab{b}})]{GPT-4}
OpenAI.
\newblock {GPT-4}.
\newblock \url{https://openai.com/gpt-4}, 2023{\natexlab{b}}.
\newblock Accessed: 2025-02-25.

\bibitem[OpenAI(2023{\natexlab{c}})]{GPT-4V(ision)}
OpenAI.
\newblock {GPT-4V(ision) System Card}.
\newblock \url{https://cdn.openai.com/papers/GPTV_System_Card.pdf}, 2023{\natexlab{c}}.
\newblock Accessed: 2025-02-26.

\bibitem[Pan et~al.(2025)Pan, Liu, Wu, Liu, Zhu, Li, Chen, Cheng, and Rueckert]{Pan2025MedVLMR1IM}
Jiazhen Pan, Che Liu, Junde Wu, Fenglin Liu, Jiayuan Zhu, Hongwei~Bran Li, Chen Chen, Ouyang Cheng, and Daniel Rueckert.
\newblock Medvlm-r1: Incentivizing medical reasoning capability of vision-language models (vlms) via reinforcement learning.
\newblock \emph{ArXiv}, 2025.

\bibitem[Plummer et~al.(2015)Plummer, Wang, Cervantes, Caicedo, Hockenmaier, and Lazebnik]{Plummer2015Flickr30kEC}
Bryan~A. Plummer, Liwei Wang, Christopher~M. Cervantes, Juan~C. Caicedo, J. Hockenmaier, and Svetlana Lazebnik.
\newblock Flickr30k entities: Collecting region-to-phrase correspondences for richer image-to-sentence models.
\newblock \emph{International Journal of Computer Vision}, 123:\penalty0 74--93, 2015.

\bibitem[Schulman et~al.(2017)Schulman, Wolski, Dhariwal, Radford, and Klimov]{PPO}
John Schulman, Filip Wolski, Prafulla Dhariwal, Alec Radford, and Oleg Klimov.
\newblock Proximal policy optimization algorithms.
\newblock \emph{arXiv:1707.06347}, 2017.

\bibitem[Shao et~al.(2024{\natexlab{a}})Shao, Qian, Xiao, Song, Zong, Wang, Liu, and Li]{Shao2024viscot}
Hao Shao, Shengju Qian, Han Xiao, Guanglu Song, Zhuofan Zong, Letian Wang, Yu Liu, and Hongsheng Li.
\newblock Visual cot: Advancing multi-modal language models with a comprehensive dataset and benchmark for chain-of-thought reasoning.
\newblock In \emph{Neural Information Processing Systems}, 2024{\natexlab{a}}.

\bibitem[Shao et~al.(2024{\natexlab{b}})Shao, Wang, Zhu, Xu, Song, Zhang, Li, Wu, and Guo]{Shao2024DeepSeekMathPT}
Zhihong Shao, Peiyi Wang, Qihao Zhu, Runxin Xu, Jun-Mei Song, Mingchuan Zhang, Y.~K. Li, Yu Wu, and Daya Guo.
\newblock Deepseekmath: Pushing the limits of mathematical reasoning in open language models.
\newblock \emph{ArXiv}, abs/2402.03300, 2024{\natexlab{b}}.

\bibitem[Sidorov et~al.(2020)Sidorov, Hu, Rohrbach, and Singh]{Sidorov2020TextCapsAD}
Oleksii Sidorov, Ronghang Hu, Marcus Rohrbach, and Amanpreet Singh.
\newblock Textcaps: a dataset for image captioning with reading comprehension.
\newblock In \emph{ECCV}, pages 742--758, 2020.

\bibitem[Singh et~al.(2019)Singh, Natarajan, Shah, Jiang, Chen, Batra, Parikh, and Rohrbach]{Singh2019TowardsVMtextvqa}
Amanpreet Singh, Vivek Natarajan, Meet Shah, Yu Jiang, Xinlei Chen, Dhruv Batra, Devi Parikh, and Marcus Rohrbach.
\newblock Towards vqa models that can read.
\newblock In \emph{IEEE/CVF Conference on Computer Vision and Pattern Recognition}, pages 8309--8318, 2019.

\bibitem[Sun et~al.(2024)Sun, An, Tian, Nan, Liu, Liu, Shah, and Chen]{Sun2024ARO}
Shilin Sun, Wenbin An, Feng Tian, Fang Nan, Qidong Liu, Jun Liu, Nazaraf Shah, and Ping Chen.
\newblock A review of multimodal explainable artificial intelligence: Past, present and future.
\newblock \emph{ArXiv}, abs/2412.14056, 2024.

\bibitem[Sutton and Barto(1998)]{Sutton1998IntroductionTR}
Richard~S. Sutton and Andrew~G. Barto.
\newblock Introduction to reinforcement learning.
\newblock \emph{ArXiv}, 1998.

\bibitem[Wang et~al.(2023)Wang, Raman, Sibue, Ma, Babkin, Kaur, Pei, Nourbakhsh, and Liu]{Wang2023DocLLMAL}
Dongsheng Wang, Natraj Raman, Mathieu Sibue, Zhiqiang Ma, Petr Babkin, Simerjot Kaur, Yulong Pei, Armineh Nourbakhsh, and Xiaomo Liu.
\newblock Docllm: A layout-aware generative language model for multimodal document understanding.
\newblock In \emph{Annual Meeting of the Association for Computational Linguistics}, 2023.

\bibitem[Wei et~al.(2024)Wei, Kong, Chen, Zhao, Ge, Yang, Sun, Han, and Zhang]{wei2024vary}
Haoran Wei, Lingyu Kong, Jinyue Chen, Liang Zhao, Zheng Ge, Jinrong Yang, Jianjian Sun, Chunrui Han, and Xiangyu Zhang.
\newblock Vary: Scaling up the vision vocabulary for large vision-language model.
\newblock In \emph{European Conference on Computer Vision}, pages 408--424. Springer, 2024.

\bibitem[Wei et~al.(2022)Wei, Wang, Schuurmans, Bosma, Chi, Xia, Le, and Zhou]{Wei2022cot}
Jason Wei, Xuezhi Wang, Dale Schuurmans, Maarten Bosma, Ed~H. Chi, F. Xia, Quoc Le, and Denny Zhou.
\newblock Chain of thought prompting elicits reasoning in large language models.
\newblock In \emph{Neural Information Processing Systems}, pages 24824--24837, 2022.

\bibitem[Wu et~al.(2024)Wu, Mao, Zhang, Xia, Dong, Cui, and Wei]{Wu2024vot}
Wenshan Wu, Shaoguang Mao, Yadong Zhang, Yan Xia, Li Dong, Lei Cui, and Furu Wei.
\newblock Mind's eye of llms: Visualization-of-thought elicits spatial reasoning in large language models.
\newblock In \emph{Neural Information Processing Systems}, 2024.

\bibitem[Ye et~al.(2022)Ye, Tian, Yan, Yang, Wang, Zhang, He, and Lin]{Ye2022ShiftingMAqrnet}
Jiabo Ye, Junfeng Tian, Ming Yan, Xiaoshan Yang, Xuwu Wang, Ji Zhang, Liang He, and Xin Lin.
\newblock Shifting more attention to visual backbone: Query-modulated refinement networks for end-to-end visual grounding.
\newblock In \emph{IEEE/CVF Conference on Computer Vision and Pattern Recognition}, pages 15481--15491, 2022.

\bibitem[Ye et~al.(2023{\natexlab{a}})Ye, Hu, Xu, Ye, Yan, Dan, Zhao, Xu, Li, Tian, Qi, Zhang, and Huang]{Ye2023mPLUGDocOwlMM}
Jiabo Ye, Anwen Hu, Haiyang Xu, Qinghao Ye, Mingshi Yan, Yuhao Dan, Chenlin Zhao, Guohai Xu, Chenliang Li, Junfeng Tian, Qiang Qi, Ji Zhang, and Feiyan Huang.
\newblock mplug-docowl: Modularized multimodal large language model for document understanding.
\newblock \emph{ArXiv}, abs/2307.02499, 2023{\natexlab{a}}.

\bibitem[Ye et~al.(2023{\natexlab{b}})Ye, Hu, Xu, Ye, Yan, Xu, Li, Tian, Qian, Zhang, et~al.]{ye2023ureader}
Jiabo Ye, Anwen Hu, Haiyang Xu, Qinghao Ye, Ming Yan, Guohai Xu, Chenliang Li, Junfeng Tian, Qi Qian, Ji Zhang, et~al.
\newblock Ureader: Universal ocr-free visually-situated language understanding with multimodal large language model.
\newblock \emph{arXiv preprint arXiv:2310.05126}, 2023{\natexlab{b}}.

\bibitem[Yu et~al.(2018)Yu, Lin, Shen, Yang, Lu, Bansal, and Berg]{Yu2018MAttNetMA}
Licheng Yu, Zhe~L. Lin, Xiaohui Shen, Jimei Yang, Xin Lu, Mohit Bansal, and Tamara~L. Berg.
\newblock Mattnet: Modular attention network for referring expression comprehension.
\newblock In \emph{IEEE/CVF Conference on Computer Vision and Pattern Recognition}, pages 1307--1315, 2018.

\bibitem[Yu et~al.(2024)Yu, Yao, Zhang, He, Han, Cui, Hu, Liu, Zheng, Sun, and Chua]{Yu2023RLHFVTT}
Tianyu Yu, Yuan Yao, Haoye Zhang, Taiwen He, Yifeng Han, Ganqu Cui, Jinyi Hu, Zhiyuan Liu, Hai-Tao Zheng, Maosong Sun, and Tat-Seng Chua.
\newblock Rlhf-v: Towards trustworthy mllms via behavior alignment from fine-grained correctional human feedback.
\newblock In \emph{IEEE/CVF Conference on Computer Vision and Pattern Recognition}, pages 13807--13816, 2024.

\bibitem[Zang et~al.(2025)Zang, Dong, Zhang, Cao, Liu, Ding, Wu, Ma, Duan, Zhang, et~al.]{zang2025internlm}
Yuhang Zang, Xiaoyi Dong, Pan Zhang, Yuhang Cao, Ziyu Liu, Shengyuan Ding, Shenxi Wu, Yubo Ma, Haodong Duan, Wenwei Zhang, et~al.
\newblock A simple yet effective multi-modal reward model.
\newblock \emph{arXiv preprint arXiv:2501.12368}, 2025.

\bibitem[Zhang(2024)]{Zhang2024ReadAT}
Jinxu Zhang.
\newblock Read and think: An efficient step-wise multimodal language model for document understanding and reasoning.
\newblock \emph{ArXiv}, 2024.

\bibitem[Zhang et~al.(2024)Zhang, Yang, Lai, Xie, and Jin]{Zhang2024DocKylinAL}
Jiaxin Zhang, Wentao Yang, Songxuan Lai, Zecheng Xie, and Lianwen Jin.
\newblock Dockylin: A large multimodal model for visual document understanding with efficient visual slimming.
\newblock \emph{ArXiv}, abs/2406.19101, 2024.

\bibitem[Zhang et~al.(2023)Zhang, Zhang, Gu, Zhou, Lipka, Yang, and Sun]{zhang2023llavar}
Yanzhe Zhang, Ruiyi Zhang, Jiuxiang Gu, Yufan Zhou, Nedim Lipka, Diyi Yang, and Tong Sun.
\newblock Llavar: Enhanced visual instruction tuning for text-rich image understanding.
\newblock \emph{arXiv preprint arXiv:2306.17107}, 2023.

\bibitem[Zhu et~al.(2015)Zhu, Groth, Bernstein, and Fei-Fei]{Zhu2015Visual7WGQ}
Yuke Zhu, Oliver Groth, Michael~S. Bernstein, and Li Fei-Fei.
\newblock Visual7w: Grounded question answering in images.
\newblock \emph{IEEE Conference on Computer Vision and Pattern Recognition}, pages 4995--5004, 2015.

\end{thebibliography}
}

\end{document}


\maketitlesupplementary

\section{Prompt Template}

As illustrated in \cref{tab:prompt_template}, the prompt template is designed to instruct the MLLM to produce structured output ${o}$, which includes both a reasoning trace and a final output encoded in designated XML-like tags (\textless think\textgreater...\textless /think\textgreater and \textless answer\textgreater...\textless /answer\textgreater).

\begin{table*}[]
\centering
\begin{tcolorbox}[title = {\textbf{ The prompt template}},
    fonttitle=\bfseries\Large,             %
    coltitle=white,                        %
    colbacktitle=deepblue,          %
    colback=contentwhite,      %
    colframe=deepblue              %
]
{
You are given an original question. Your task is to provide an accurate answer to the question and determine the bounding box coordinates of the region that best supports your answer.\\

To enhance clarity and interpretability, you should:\\
- Understand the intent behind the original question.\\
- Modify the original question by adding relevant descriptive phrases and details based on the provided image.\\
- Ensure that the modified question remains semantically similar to the original.\\

Your response has two parts:\\
1. **Thinking Process:** Before outputting the answer, describe your reasoning process within \textless think\textgreater\textless /think\textgreater tags.\\
2. **Final Output:** Provide the answer in JSON format within \textless answer\textgreater\textless /answer\textgreater tags. The JSON should contain the following keys:\\
\hspace*{1em} - **rephrase\_question**: The improved and more descriptive version of the original question.\\
\hspace*{1em} - **bbox\_2d**: The bounding box coordinates [x\_min, y\_min, x\_max, y\_max] of the region that supports the answer.\\
\hspace*{1em}  - **final\_answer**: The actual answer to the question.\\

\#\#\# **Example Output Format:**\\
\#\#\# Original question: ``What is the man doing?"\\
\textless think\textgreater\\
reasoning process here \\
\textless /think\textgreater\\
\textless answer\textgreater\\
\{\\
\hspace*{1em}     ``rephrase\_question": ``What is the man wearing while preparing to shoot the basketball near the hoop?",\\
\hspace*{1em}     ``bbox\_2d": [150, 300, 400, 600],\\
\hspace*{1em}     ``final\_answer": ``answer here."\\
\}\\
\textless /answer\textgreater\\

\#\#\# Original question: ``\{\textcolor{red}{Question}\}"\\
}
\end{tcolorbox}
\caption{\textbf{The template of our employed prompt for \ourmodel.} \textcolor{red}{Question} will be replaced with the specific question during training and inference.}
\label{tab:prompt_template}
\end{table*}

\section{Rewiew of GRPO}
The Group Relative Policy Optimization (GRPO) algorithm, first introduced in DeepSeekMath~\cite{Shao2024DeepSeekMathPT}, is a reinforcement learning framework designed to improve reasoning without the need for a separate critic model, a key limitation of existing methods such as Proximal Policy Optimization (PPO)\cite{PPO}. Traditional RL approaches like PPO rely on a value network to estimate the quality of model predictions, which can introduce instability and additional computational costs. In contrast, GRPO directly compares a group of generated responses, making it a more efficient alternative for large-scale language model training.

In GRPO, given a question $q$, the old policy model $\pi_{\theta_{\text{old}}}$ first generates a group of different candidate response outputs $\{o_1, o_2, ..., o_G\}$ with size of $G$. These response outputs are then evaluated through a rule-based reward function $R(q, o)$ to obtain $G$ rewards denoted as $\{r_1, r_2, ..., r_G\}$ correspondingly, which is defined as follows:
\begin{equation}
    r_i = R(q, o_i) = 
    \begin{cases} 
    1 ,& \text{if } o_i \text{ = ground truth}, \\
    0 ,& \text{otherwise}.
    \end{cases} 
\end{equation}
where $R(\cdot, \cdot)$ is the rule-based verifiable reward function. $R$ takes the question and output pair $(q,o_i)$ as inputs, and checks whether the prediction $o_i$ is correct compared to ground truth under predefined rules. In our works, we proposed multi-objective reward functions tailored for document understanding, to incentivize the model to generate human-understandable reasoning steps, while ensuring robust generalization across diverse document types and tasks.

Instead of computing absolute values for each response, GRPO normalizes the rewards within the group, ensuring that the model learns from relative advantages. Specifically, the advantage is computed by taking the difference between each reward and the $mean$ of the group, normalized by the standard deviation $std$, formulated as follows: 
\begin{align}
A_i &= \frac{r_i - \text{mean}(\{r_1, \dots, r_G\})}{\text{std}(\{r_1, \dots, r_G\})},
\end{align}
where $A_i$ represents the advantage of $i$-th output $o_i$, meaning the relative quality of the $i$-th responses. The advantage $A_i$ is sequence-level normalized reward, and we set the advantage $A_{i,t}$ of $t$-th auto-regressive decoding time step token in the output $o_i$ as the sequence-level advantage $A_i$. This process eliminates the need for a critic network, making policy updates computationally efficient and stable. The intuition behind GRPO objective is to maximize the advantage of the generated responses, while ensuring that the model remains close to the reference policy model $\pi_{{ref}}$. Consequently, the GRPO loss $\mathcal{L}_{\text{GRPO}}$ is defined as follows: 
\begin{align}
\mathcal{L}_{\text{GRPO}}(\theta) &= -\frac{1}{G}\sum_{i=1}^G \frac{1}{\left|o_i\right|} \sum_{t=1}^{\left|o_i\right|}  \bigg[  
\frac{\pi_{\theta}(o_{i,t} \mid q, o_{i,\textless t})}{ \varphi\big[\pi_{\theta}(o_{i,t} \mid q, o_{i,\textless t} ) \big] } A_{i,t}  \notag \\
&\quad - \beta \mathbb{D}_{\text{KL}}(\pi_{\theta} \parallel \pi_{\text{ref}}) \bigg] ,
\label{eq:simply_grpo}
\end{align}
where the first term represents the scaled advantage and the second term is regularization to penalize deviations from the reference policy $\pi_{ref}$ through Kullback–Leibler (KL) divergence $\mathbb{D}_{\text{KL}}(\pi_{\theta} \parallel \pi_{\text{ref}})$, helping prevent catastrophic forgetting. $\varphi[\cdot]$ represents stop gradient operation. $\theta$ is the trainable parameter of the current policy model $\pi_{\theta}$. $\beta \in \mathbb{R} \geq 0$ is a hyper-parameter and controls the regularization strengths. GRPO encourages the model to favor better answers with a high reward value within the group.

In the original DeepSeekMath~\cite{Shao2024DeepSeekMathPT} paper, the objective $\mathcal{L}_{\text{GRPO}}$ formulation in \cref{eq:simply_grpo} is generalized to account for multiple updates after each group response generation by leveraging the clipped surrogate objective to ensure that updates do not deviate excessively from the reference policy by bounding the policy ratio between $1-\epsilon$ and $1+\epsilon$ via $\text{clip}(\cdot, 1-\epsilon, 1+\epsilon)$ function, formulated as follows:
\begin{align}
\mathcal{L}_{\text{GRPO}}(\theta) &= -\frac{1}{G}\sum_{i=1}^G \frac{1}{\left|o_i\right|} \sum_{t=1}^{\left|o_i\right|} \bigg[ \min \bigg(  
\frac{\pi_{\theta}(o_{i,t} \mid q, o_{i,\textless t})}{\pi_{\theta_{\text{old}}}(o_{i,t} \mid q, o_{i,\textless t} )} A_{i,t},  \notag \\
&\quad \text{clip} \left( \frac{\pi_{\theta}(o_{i,t} \mid q,o_{i, \textless t})}{\pi_{\theta_{\text{old}}}(o_{i,t} \mid q, o_{i, \textless t})}, 1 - \epsilon, 1 + \epsilon \right) A_{i,t} \bigg) \notag \\
&\quad - \beta \mathbb{D}_{\text{KL}}(\pi_{\theta} \parallel \pi_{\text{ref}}) \bigg] ,
\label{eq:grpo}
\end{align}
where $\epsilon \in \mathbb{R} \geq 0$ is a clipping-related hyper-parameter introduced in PPO~\cite{PPO} for stabilizing training by preventing drastic changes in policy updates. In practice, as in the original paper, we only do one update per generation. In this condition, $\pi_{\theta}$ is equal to $\pi_{\theta_{odd}}$, so we can simplify the loss to the first form defined in \cref{eq:simply_grpo}.

In practice, KL divergence is estimated using the unbiased estimator introduced by~\cite{Schulmankl}. The approximator is defined as follows:
\begin{align}
\mathbb{D}_{\mathrm{KL}}\left[\pi_\theta \| \pi_{\mathrm{ref}}\right] &=\frac{\pi_{\mathrm{ref}}\left(o_{i, t} \mid q, o_{i,<t}\right)}{\pi_\theta\left(o_{i, t} \mid q, o_{i,<t}\right)} \notag \\
&\quad -\log \frac{\pi_{\mathrm{ref}}\left(o_{i, t} \mid q, o_{i,<t}\right)}{\pi_\theta\left(o_{i, t} \mid q, o_{i,<t}\right)}-1,
\end{align}
where this approximation ensures that KL estimates remain positive and computationally stable throughout training.

\begin{table*}[t]
\centering

 \begin{minipage}[b]{0.72\textwidth}
 \resizebox{1\linewidth}{!}{%
\begin{tabular}{l|c|c|c|c|c|c|c|c}
\toprule
\multicolumn{9}{c}{{Document-oriented Understanding }}  \\
\midrule
  \multicolumn{3}{c|}{} & \multicolumn{5}{c|}{{Doc/Text}}          & {Chart}      \\
 \cmidrule(lr){1-3}\cmidrule(lr){4-8} \cmidrule(lr){9-9} 
MLLM & Res. & Strategy & DocVQA & TextCaps & TextVQA & \cellcolor{lightgrey} DUDE & \cellcolor{lightgrey} SROIE & InfoVQA \\
\midrule

VisCoT-7B~\cite{Shao2024viscot} &   224$^2$ & SFT & 13.6   &  41.3  & 46.8  &  \cellcolor{lightgrey}5.0    &  \cellcolor{lightgrey}15.7  &  7.2    \\
VisCoT-7B~\cite{Shao2024viscot}  &   336$^2$  & SFT & 20.4 & 46.3 & 57.6 & \cellcolor{lightgrey}9.6 & \cellcolor{lightgrey}18.5 & 10.0      \\
\midrule
\ourmodel-7B  &   336$^2$ & RL  & \textbf{38.3} & \textbf{58.6} & \textbf{59.2 }& \cellcolor{lightgrey} \textbf{27.5} & \cellcolor{lightgrey}\textbf{32.1} & \textbf{23.6  }    \\
 \bottomrule
\end{tabular}
}
\end{minipage}
 
 \begin{minipage}[b]{0.72\textwidth}
 \resizebox{1\linewidth}{!}{%
\begin{tabular}{l|c|c|c|c|c|c|c|c}
\multicolumn{9}{c}{{General Multimodal Understanding }}  \\
\midrule
  \multicolumn{3}{c|}{}      & \multicolumn{2}{c|}{{General VQA}} & \multicolumn{3}{c|}{{Relation Reasoning}}   &   \multirow{2}{*}{{Average}}  \\
 \cmidrule(lr){1-3}\cmidrule(lr){4-5} \cmidrule(lr){6-8} 
MLLM & Res. & Strategy & Flickr30k   & \cellcolor{lightgrey}Visual7W & GQA      & Open Images      & VSR      &  \\
\midrule

VisCoT-7B~\cite{Shao2024viscot} &   224$^2$   & SFT  &   49.6  & \cellcolor{lightgrey}31.1 &  42.0  & 57.6  &  69.6    &   37.2  \\
VisCoT-7B~\cite{Shao2024viscot}  &   336$^2$  & SFT  & 51.3 & \cellcolor{lightgrey}29.4 & 49.5 & 59.3 & 54.0  & 37.6 \\
\midrule
\ourmodel-7B  &  336$^2$  & RL  & \cellcolor{lightgrey}\textbf{55.7} & \cellcolor{lightgrey}\textbf{36.3} & \cellcolor{lightgrey}\textbf{53.6} & \cellcolor{lightgrey}\textbf{67.1} & \cellcolor{lightgrey}\textbf{59.8}  & \cellcolor{lightgrey}\textbf{46.5} \\
 \bottomrule
\end{tabular}
}
\end{minipage}

\caption{Detection performance (Top-1 Accuracy@0.5) on the Visual CoT benchmark~\cite{Shao2024viscot}. \colorbox{lightgrey}{Grey} results indicate zero-shot performance. Res. shorts for image resolution. Average refers to the average accuracy across eleven datasets. The ground truth bounding boxes used for computing the metric are the intermediate CoT bounding boxes annotated in the Visual CoT benchmark.}
\label{table:supp_det_cot_benchmark}

\end{table*}

\begin{table*}[t]
\centering

\resizebox{0.97\linewidth}{!}{\begin{tabular}{c|cc|ccc|ccc}
\toprule      \multirow{2}{*}{Method}
             & \multicolumn{2}{c|}{Scene Text-Centric VQA}        & \multicolumn{3}{c|}{Document-oriented VQA}                    & \multicolumn{3}{c}{KIE}   \\
              \cmidrule(lr){2-3}\cmidrule(lr){4-6} \cmidrule(lr){7-9} 
             & STVQA & TextVQA & DocVQA & InfoVQA & ChartQA & FUNSD   & SROIE  & POIE  \\ \midrule
BLIP2-OPT-6.7B~\cite{Li2023BLIP2BL}   & 20.9  & 23.5  & 3.2    & 11.3           & 3.4                  & 0.2     & 0.1    & 0.3  \\
mPLUG-Owl~\cite{Ye2023mPLUGOwlME}    & 30.5  & 34.0  & 7.4    & 20.0             & 7.9           & 0.5     & 1.7    & 2.5   \\
InstructBLIP~\cite{Dai2023InstructBLIPTG} & 27.4  & 29.1   & 4.5    & 16.4           & 5.3             & 0.2     & 0.6    & 1.0     \\
LLaVAR~\cite{zhang2023llavar}       & 39.2  & 41.8   & 12.3   & 16.5           & 12.2              & 0.5     & 5.2    & 5.9    \\
BLIVA~\cite{Hu2023BLIVAAS}        & 32.1  & 33.3  & 5.8    & 23.6           & 8.7             & 0.2     & 0.7    & 2.1     \\
mPLUG-Owl2-8~\cite{Ye2023mPLUGOwI2RM}   & 49.8  & 53.9   & 17.9   & 18.9           & 19.4            & 1.4     & 3.2    & 9.9         \\
LLaVA1.5-7B~\cite{liu2024llavanextllava1.5}     & 38.1  & 38.7    & 8.5    & 14.7           & 9.3            & 0.2     & 1.7    & 2.5          \\
TGDoc~\cite{Wang2023TowardsIDtgdoc}   & 36.3 & 46.2  & 9.0           & 12.8                & 12.7       & 1.4    & 3.0   & 22.2   \\
UniDoc~\cite{Feng2023UniDocAU}       & 35.2  & 46.2    & 7.7    & 14.7           & 10.9                    & 1.0       & 2.9    & 5.1    \\
DocPedia~\cite{feng2024docpedia}     & 45.5  & 60.2   & 47.1   & 15.2           & 46.9                 & {29.9}    & 21.4   & 39.9    \\
Monkey-8B~\cite{Li2023MonkeyIR}       & {54.7}  & 64.3    & {50.1}   & {25.8}           & {54.0}            & 24.1    & {41.9}   & 19.9  \\ 
InternVL-8B~\cite{Chen2023InternVLS}     & 62.2  & 59.8      &28.7    & 23.6           & 45.6  & 6.5    & 26.4    & 25.9   \\
InternLM-XComposer2-7B~\cite{Dong2024InternLMXComposer2MF}      & 59.6 & 62.2      &39.7    & 28.6           & 51.6  & 15.3   & 34.2    & 49.3   \\ 
TextMonkey-9B~\cite{Liu2024TextMonkeyAO}   & 61.8 & 65.9    & 64.3          & 28.2                & 58.2       & 32.3    & 47.0   & 27.9   \\
InternVL2-2B~\cite{Chen2024HowFAinternvl2}   & 65.6 & 66.2    & 76.7          & 46.8                & {67.6}       & 42.0    & 68.0   & 66.8   \\
Mini-Monkey-2B~\cite{Huang2024MiniMonkeyMA} & {67.2} & {68.8} & {78.4} & {50.0} & 67.3 & {43.2} & {70.5} & {71.2} \\ 
\midrule
\ourmodel-7B & \textbf{68.4} & \textbf{69.7} & \textbf{78.8} & \textbf{52.3} & \textbf{67.8 }& \textbf{47.2} & \textbf{73.1} & \textbf{72.8} \\ 
\bottomrule
\end{tabular}}
\caption{Quantitative accuracy (\%) comparison of \ourmodel with existing multimodal large language models (MLLMs) on widely used benchmark. Following TextMonkey~\cite{Liu2024TextMonkeyAO}, we use the accuracy metrics to evaluate our method.}
\label{tab:supp_ocrbench_accuracy_metric}
\end{table*}

\section{More Results and Analysis}

\mypara{RoI Detection Results.}
\cref{table:supp_det_cot_benchmark} presents the RoI detection performance, measured by Top-1 Accuracy@0.5, across multiple document understanding and general reasoning tasks. A higher score indicates better alignment between the model’s predicted bounding boxes and the ground truth key regions annotated in the Visual CoT benchmark~\cite{Shao2024viscot}.
Compared to VisCoT-7B, our model \ourmodel-7B (336²) achieves substantial improvements across all tasks, particularly in document-oriented datasets. It outperforms the strongest baseline by a large margin on DocVQA (38.3 vs. 20.4), TextCaps (58.6 vs. 46.3), and TextVQA (59.2 vs. 57.6). More challenging datasets, such as DUDE and SROIE, which require precise text-region localization, also see significant gains, with our model scoring 27.5 and 32.1, compared to 9.6 and 18.5, respectively.
Beyond document tasks, \ourmodel demonstrates stronger generalization in VQA and relational reasoning benchmarks, outperforming VisCoT-7B in Flickr30k (55.7 vs. 51.3), GQA (53.6 vs. 49.5), and Open Images (67.1 vs. 59.3). The model also improves Visual7W and VSR performance, achieving 36.3 and 59.8, respectively. These results confirm that reinforcement learning with RoI-based rewards enhances the model’s ability to precisely localize key regions, leading to better multimodal alignment and more reliable reasoning outputs.
The superior results demonstrate \ourmodel's effectiveness in both structured document reasoning and general multimodal comprehension.

\mypara{OCRBench Results.}
To further evaluate \ourmodel beyond the Visual CoT Benchmark~\cite{Shao2024viscot}, we assess its performance on OCRBench~\cite{Liu2023OCRBenchOT}, a widely used benchmark for text-centric multimodal understanding. Following the TextMonkey~\cite{Liu2024TextMonkeyAO} evaluation framework, we use accuracy metrics (\%) across scene text-based VQA, document-oriented VQA, and key information extraction (KIE) tasks.
As shown in \cref{tab:supp_ocrbench_accuracy_metric}, \ourmodel-7B achieves state-of-the-art performance, surpassing previous MLLMs across all categories. In scene text VQA, our model scores 68.4\% on STVQA and 69.7\% on TextVQA, outperforming Mini-Monkey-2B~\cite{Huang2024MiniMonkeyMA} and InternVL2-2B~\cite{Chen2024HowFAinternvl2}. In document-oriented VQA, \ourmodel reaches 78.8\% on DocVQA, 52.3\% on InfoVQA, and 67.8\% on ChartQA, consistently leading across structured text understanding tasks.
For key information extraction (KIE), which demands precise text localization and recognition, \ourmodel sets new benchmarks with 47.2\% on FUNSD, 73.1\% on SROIE, and 72.8\% on POIE, surpassing Mini-Monkey-2B and other strong baselines. These results highlight the effectiveness of reinforcement learning with structured rewards in improving both text-centric reasoning and document comprehension, demonstrating \ourmodel's ability to generalize across complex multimodal text understanding tasks.

\mypara{Accuracy of Rephrased Questions.} We construct a new training set using model generated rephrased questions and fine-tuned on Qwen. As shown in \cref{tab:acc_rephrase_ques}, this model outperforms one trained on original QA pairs (0.548 vs. 0.497 average score on Visual CoT), demonstrating that rephrased questions preserve and even enhance task relevance.

\begin{table}[h]
\centering
\scalebox{.8}{
\begin{tabular}{llll}
\toprule
Method                            & Res.                  & Data                     & Avg. \\
\midrule
\multirow{2}{*}{Qwen2.5VL-7B} & \multirow{2}{*}{$336^2$} & Original   QA & 0.497   \\
                                  &                       & Rehprase   QA & \textbf{0.548}  \\
\bottomrule
\end{tabular}}
\caption{Accuracy of rephrased questions.}
\label{tab:acc_rephrase_ques}
\end{table}

\mypara{Hallucination in Rephrased Questions.} Following HallusionBench~\cite{Liu2023HallusionBenchYS}, we use GPT-4 to judge 200 randomly sampled rephrased questions. As shown in \cref{tab:hallucination_rephrase_ques}, results show 96\% correctness, 0\% inconsistency, and 4\% unclear, indicating that language hallucinations are rare. Besides, a human evaluation confirms 99\% correctness.

\begin{table}[h]
\centering
\scalebox{.8}{
  \begin{tabular}{@{}lcccc@{}}
    \toprule
    \multirow{2}{*}{} & \multicolumn{3}{c}{Semantic   Consistency} & \multirow{2}{*}{\makecell[l]{Human \\Check}} \\
    \cmidrule(lr){2-4}
                                   & Correct     & Inconsistent    & Unclear    \\
    \midrule
    \makecell[l]{Rephrase \\ question}            & 96\%          & 0\%               & 4\% & 99\% \\
    \bottomrule
    \end{tabular}}
\caption{Hallucination in rephrased questions.}
\label{tab:hallucination_rephrase_ques}
\end{table}

\mypara{Hallucination of the Resulting Model.} 
 We evaluated hallucination rate of the resulting model on HallusionBench~\cite{Liu2023HallusionBenchYS}. As shown in \cref{tab:hallucination_res_model}, our model achieves 69.8\%, outperforming baseline Qwen2.5VL (69.4\%).

\begin{table}[h]
\centering
\scalebox{.8}{
  \begin{tabular}{@{}lc@{}}
    \toprule
Method                            & HallusionBench           \\
\midrule
Qwen2.5VL-7B & 69.4\%   \\
Ours         & \textbf{69.8\%}  \\
  \bottomrule
  \end{tabular}}
\caption{Hallucination of the resulting model.}
\label{tab:hallucination_res_model}
\end{table}

\begin{figure}[htbp]
  \centering
  \includegraphics[width=0.28\textwidth]{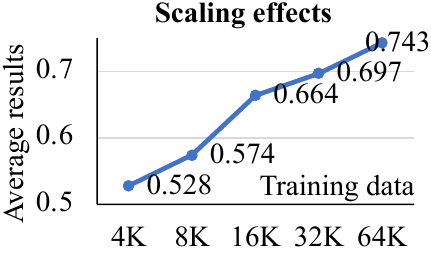}
  \caption{Scaling effects of DocThinker on document benchmarks. Increasing training data from 4k to 64k consistently improves performance, showing the scalability of GRPO.}
  \label{fig:scaling_effects}
\end{figure}

\mypara{Scaling Effects.} We scale training data from 4k to 64k using samples from Visual CoT. As shown in \cref{fig:scaling_effects}, the average results of DocThinker-7B ($336^2$) improve consistently with more data, demonstrating a clear scaling effect. 

\noindent 

\begin{figure}[htbp]
  \centering
  \includegraphics[width=0.5\textwidth]{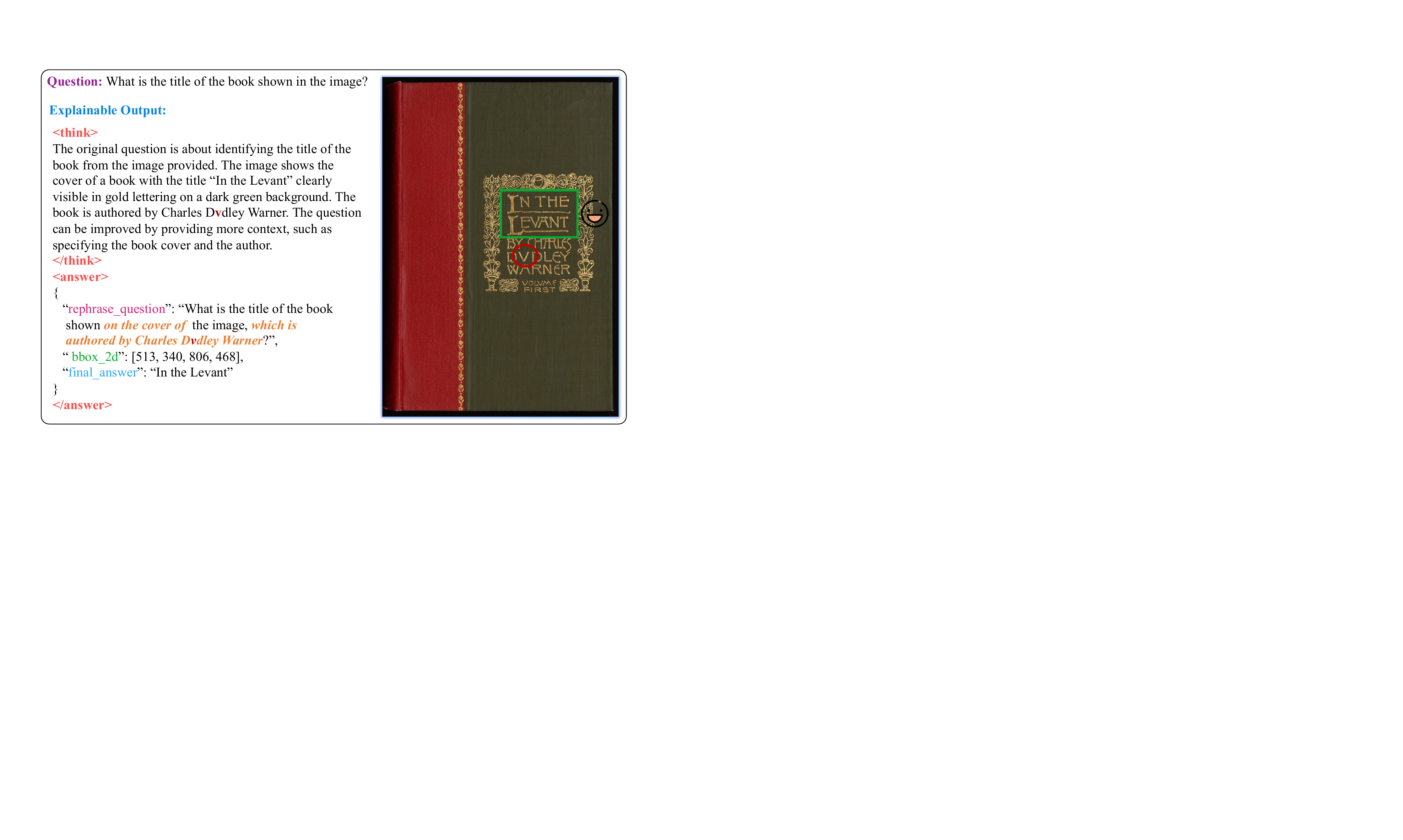}
  \caption{Failure case. Although the model correctly predicts the final answer ``In the Levant'' and localizes the 2D bounding box of RoI accurately, the rephrased question contains a hallucination: it misidentifies the author as ``Charles D\textcolor{red}{v}dley Warner'' instead of ``Charles D\textcolor{red}{u}dley Warner''. This error likely results from visual confusion caused by the stylized font on the book cover, where the letter ``\textcolor{red}{u}'' was misread as ``\textcolor{red}{v}''. While the final output remains correct, this case highlights the model’s vulnerability to artistic distortions in text recognition.}
  \label{fig:failure_case}
\end{figure}

\mypara{Failure Case.}
As shown in \cref{fig:failure_case}, although the model correctly locates the 2D bounding box of RoI and predicts the final answer ``In the Levant'', the rephrased question incorrectly identifies the author as ``Charles D\textcolor{red}{v}dley Warner'' instead of the correct ``Charles D\textcolor{red}{u}dley Warner''. This error likely stems from visual confusion caused by the stylized font on the book cover, where the letter ``\textcolor{red}{u}'' was misperceived as ``\textcolor{red}{v}''. While this hallucination does not affect the final answer, it reveals a limitation in the model's visual text recognition. Future work may improve robustness against such artistic distortions to reduce semantic hallucinations in intermediate outputs.

{
    \small
    \bibliographystyle{ieeenat_fullname}
    \bibliography{main}
}